\documentclass{article}

\usepackage{arxiv}
\usepackage{sidecap}
\usepackage[utf8]{inputenc} 
\usepackage[T1]{fontenc}    
\usepackage{hyperref}       
\usepackage{url}            
\usepackage{booktabs}       
\usepackage{amsfonts}       
\usepackage{amsmath}       
\usepackage{nicefrac}       
\usepackage{microtype}      
\usepackage{lipsum}
\usepackage{graphicx}
\usepackage{subfigure}
\usepackage[font=small]{caption}
\usepackage{multicol}
\usepackage{multirow}
\usepackage{afterpage}
\usepackage{xcolor}
\usepackage{microtype}
\graphicspath{ {./images/} }

\title{Unifying Cardiovascular Modelling with Deep Reinforcement Learning for Uncertainty Aware Control of Sepsis Treatment}

\author{
 Thesath Nanayakkara \\
  Department of Mathematics\\
  University of Pittsburgh\\
  Pittsburgh, PA, 15213 \\
  
  \And
Gilles Clermont\\
Department of Critical Care Medicine\\ 
The Clinical Research, Investigation, and Systems Modeling of Acute Illness (CRISMA) Center\\
University of Pittsburgh School of Medicine\\
Pittsburgh, PA, 15213\\
  \And
Christopher James Langmead\\
Computational Biology Department\\
School of Computer Science\\
Carnegie Mellon University\\
Pittsburgh, PA, 15213\\

\And
David Swigon\\
Department of Mathematics\\ 
McGowan Institute for Regenerative Medicine\\
University of Pittsburgh\\
Pittsburgh, PA, 15213\\

}

\begin{document}

\maketitle

\vspace{20pt}

\begin{abstract}

{Sepsis is a potentially life-threatening inflammatory response to infection or severe tissue damage. 
It has a highly variable clinical course, requiring constant monitoring of the patient’s state to guide
the management of intravenous fluids and vasopressors, among other interventions. Despite decades of research, there's still debate among experts on optimal treatment.
Here, we combine for the first time, distributional deep reinforcement learning with mechanistic physiological models to find personalized sepsis treatment strategies. Our method handles partial observability by leveraging known cardiovascular physiology, introducing a novel physiology-driven recurrent autoencoder, and quantifies the uncertainty of its own results. Moreover, we introduce a framework for uncertainty-aware decision support with humans in the loop. We show that our method learns physiologically explainable, robust policies, that are consistent with clinical knowledge. Further our method consistently identifies high-risk states that lead to death, which could \emph{potentially} benefit from more frequent vasopressor administration, providing valuable guidance for future research.}

\footnote{Code for this work can be found at //github.com/thxsxth/POMDP\_RLSepsis}
\end{abstract}


Sepsis is a major host response to infection which can result in tissue damage, organ damage and death. The mortality and economic burden of sepsis is very large. In the U.S., sepsis is responsible for 6\% of all hospitalizations and 35\% of all in-hospital deaths \cite{liu2014hospital,rhee2017incidence}, and an economic burden of more than \$20B per year \cite{paoli_reynolds_sinha_gitlin_crouser_2018}. The treatment of sepsis is extremely challenging, due to the high variability among patients, with respect to both the progression of the disease, the host response to infection, and the response to medical interventions, suggesting the need for a dynamic and personalized approach to treatment \cite{marik2015demise,lazuar2019precision,douglas2020fluid}. Presently, the search for treatment strategies to optimize sepsis patient outcomes remains an open challenge in critical care medicine, despite decades of research.  

Recently, there has been considerable interest in the application of  Reinforcement Learning (RL) \cite{sutton1998rli} to extract vasopressor and intravenous (IV) fluid treatment policies (i.e., strategies) for septic patients from electronic health records data (ex. \cite{komorowski2018artificial,raghu2017deep,peng2018improving,li2019optimizing,killian2020empirical}). Informally, the goal is to learn a policy that maps the patient's current state to an action (i.e., medical intervention), so as to maximize the chances of future recovery. The RL framework is well-matched to the actual behaviors of physicians, who continuously observe, interpret, and react to their patient's condition. The promise of RL in medicine is that we \emph{might} be able to find policies that outperform humans (as it has in other domains, ex. \cite{mnih2015human,silver2016mastering,fuchs2020super}), by automatically personalizing the treatment strategy for each patient, as opposed to using one that is expected to work well on the \emph{typical} patient \cite{liu2020reinforcement,yu2019reinforcement}. However, there are many challenges that must be met before RL can be used to guide medical decision making in real-life settings \cite{gottesman2019guidelines}.

A particularly severe challenge is partial observability of patient state. Despite the richness of data collected at the ICU, the mapping between true patient states and clinical observables is often ambiguous. We believe that this ambiguity can be reduced through the use of mechanistic mathematical models of physiology that relate observables to a more complete representation of the patient's cardiovascular state. Such models are plentiful in the literature, and embody decades of research in physiology and medicine. Our proposed solution integrates, for the first time, a clinically relevant mechanistic model into a Deep RL framework. The specific model we use was chosen because it estimates the unobservable aspects of cardiovascular state that are relevant to specific interventions (vasopressors and IV fluids), and the clinician's goals --- counteracting hypovolemia, vasodilation, and other physiological disturbances. This model is integrated into our framework using a self-trained deep recurrent autoencoder that uses a variety of inputs, including the patient's vital signs, organ function scores, and previous treatments.

The second challenge addressed by our framework is uncertainty in the learned policy, and thus the expected outcomes. Similar to previous efforts to extract sepsis treatment policies from retrospective data (ex. \cite{komorowski2018artificial}), our method works in the Batch Reinforcement Learning setting \cite{lange2012batch}, where the agent cannot explore the environment freely. In this setting, it is well known that RL can perform poorly \cite{fujimoto2019off}, if the agent encounters states that are rare or even unobserved in the training data. For this reason, it has been argued that all forms of uncertainty should be quantified in any application of Artificial Intelligence to Medicine \cite{osti_1561669}. Thus, we quantify model uncertainty\footnote{Here, and throughout this work we use the term 'model uncertainty' to mean the uncertainty in the outputs of neural networks used for RL. This should be not be confused with the model-based vs model-free RL distinction, because once we have inferred latent states, our approach qualifies as 'model-free'. The literature also uses the term epistemic uncertainty and parametric uncertainty for model uncertainty.} via bootstrapping and take a distributional approach to factor in environment uncertainty. We also propose a decision framework where the clinician is presented with a quantitative assessment of the distribution over outcomes for each state-action pair.

\section{Background and Related work}
\label{sec:headings}

\subsection*{Reinforcement Learning}
Reinforcement Learning is a framework for optimizing sequential decision making. In its standard form, a Markov Decision Process (MDP), consisting of a 5-tuple ($S$,$A$,$r$,$\gamma$,$p$) is the framework considered. Here, $S$ and $A$ are state and action spaces, $r:(S,A,S) \to \mathbb{R}$ is a reward function, $p :(S,A,S) \to [0,\infty)$ denotes the unknown environment dynamics, which specifies the distribution of the next state $s'$, given the state-action pair $(s,a)$, and $\gamma$ is a discount rate applied to rewards. A policy is (a possibly stochastic) mapping from $S$ to $A$. The agent aims to compute the policy $\pi$ which maximizes the expected future reward $E_{p,\pi}[\Sigma_{t}\gamma^{t}r_{t}]$. In the partially observed setting there is a distinction between the observations, denoted as $o_{t}$, and the state $s_{t}$, and the environment dynamics includes the conditional probability density $p(o_{t}|s_{t})$. This extends the MDP formalism to that of Partially Observed Markov Decision Process (POMDP).


The search for of an optimal policy can be performed in several ways, including the iterative calculation of the \emph{value function},
$V^{\pi}(s)=\mathbb{E}_{p,\pi}[\Sigma_{t}\gamma^{t}r_{t}(s_t,a_t)|s_{0}=s,\pi], \forall s\in S$, which returns the expected future discounted rewards when following policy $\pi$ and starting from the state $s$, or the \emph{Q-function}, $Q^{\pi}(s,a)=\mathbb{E}_{p,\pi}[\Sigma_{t}\gamma^{t}r_{t}(s_t,a_t)|s_{0}=s,\pi,a_{0}=a], \forall s\in S, a\in A$, which returns the expected future reward when choosing action $a$ in state $s$, and then following policy $\pi$. Central to many RL algorithms is the Bellman equation \cite{bellman1965dynamic}:

\begin{equation}
 Q^{\pi}(s,a)=\mathbb{E}_{p}[r(s,a)]+\gamma \mathbb{E}_{p,\pi}[Q^{\pi}(s',a')],
 \label{eq:bell1}
\end{equation}

and the Bellman optimality equation:

\begin{equation}
 Q^{*}(s,a)=\mathbb{E}_{p}[r(s,a)]+\gamma \mathbb{E}_{p}[\max_{a'\in A}Q^{*}(s',a')]
 \label{eq:bell_2}
\end{equation} (where $Q^{*}(s,a)$ is the optimal $Q$ function, and $s'$ denotes the random next state).

\begin{figure}[ht]
\centering
\includegraphics[scale=0.16]{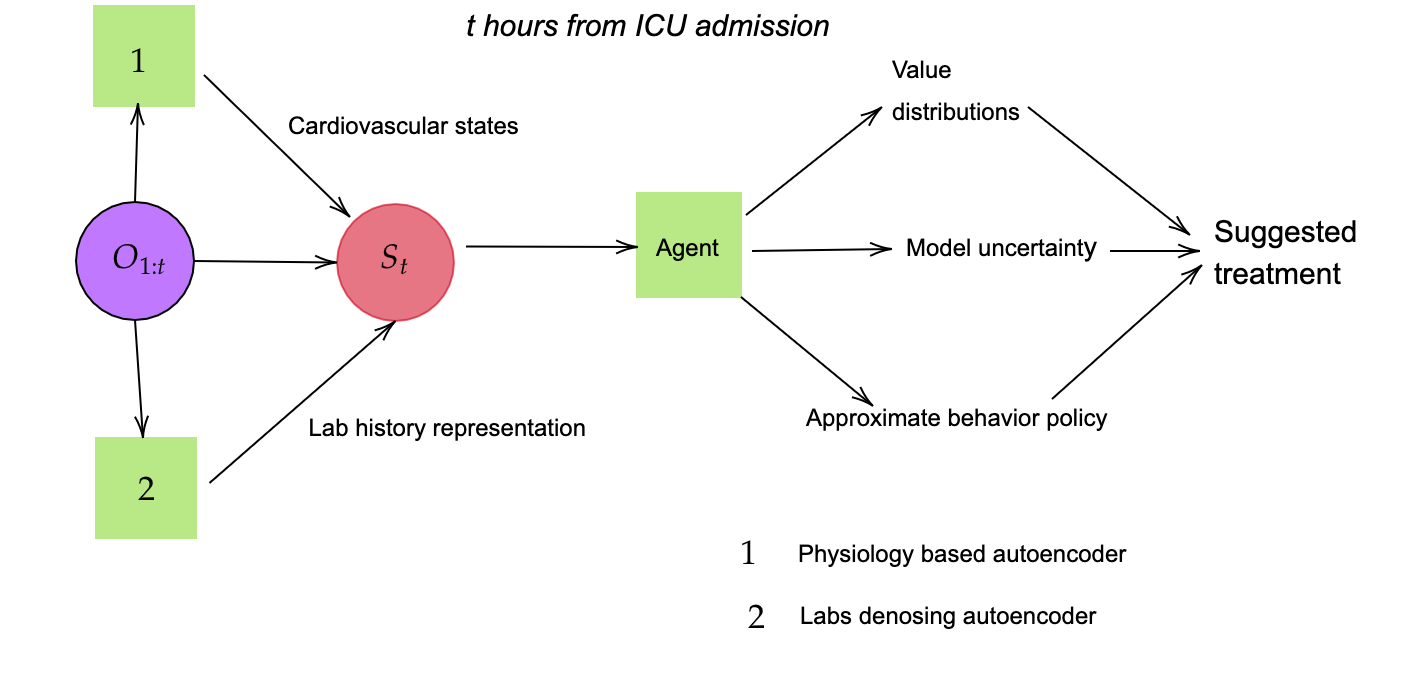}{a}
\hspace{15pt}    
\includegraphics[scale=0.4]{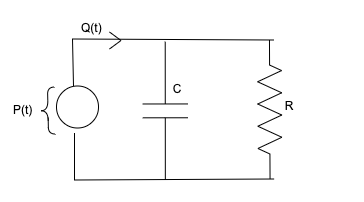}{b}
~

\includegraphics[width=200 pt]{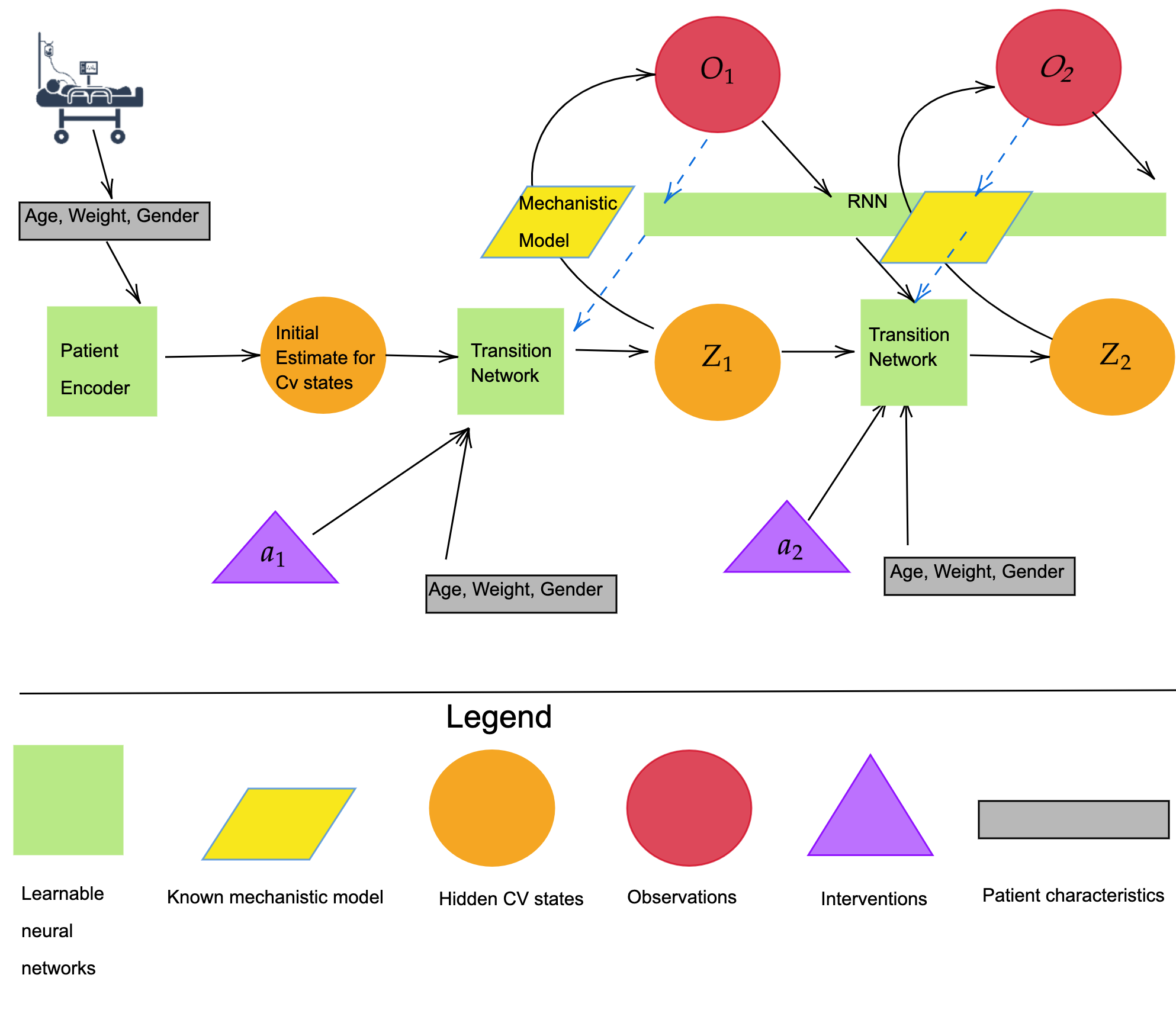}{c}
\hspace{15pt}
\caption{\textbf{Proposed decision support system (a)}: We use the compete patient history, which includes, vitals, scores, and labs, and previous treatment, to infer hidden states. These would all combine to make the state $S_t$. Our trained agent, takes this state and outputs value distributions for each treatment, it's own uncertainty, and an approximate clinician's policy. We then factor in all 3 to propose uncertainty-aware treatment strategies. \textbf{The electrical analog of the cardiovascular model (b)} This provides a lumped representation of the resistive and elastic properties of the entire arterial circulation using just two elements, a resistance R and a capacitance C. This model is used to derive algebraic equations relating R, C, stroke volume (SV), filling time (T), to heart rate (F) and pressure. The Cardiac Output (CO) can be then computed as (SV)F. These equations define the decoder of the physiology-driven autoencoder.
\textbf{Complete physiology-driven autoencoder network structure (c)} 
Patient history is sequentially encoded using three neural networks. A patient encoder computes initial cardiovascular state estimates using patient characteristics, a recurrent neural network (RNN) encodes the past history of vitals and scores, up to and including the current time point, and a transition network which takes the previous cardiovascular state, the action and the history representation to output new cardiovascular state estimates.
}
\label{fig:overall}
\end{figure}

\subsection*{Distributional and Uncertainty Aware Reinforcement Learning}
Distributional Reinforcement Learning \cite{pmlr-v70-bellemare17a,pmlr-v84-rowland18a,barth2018distributed} extends traditional RL methods by estimating the entire return distribution from a given state, rather than simply an expected value. It has been shown that distributional RL can achieve superior performance in the context of Batch RL \cite{agarwal2020optimistic}. For this reason, and because distributions are relevant to our overall goal of providing clinicians with an assessment of the range of possible outcomes for each state-action pair, we employ Categorical Distributional RL \cite{pmlr-v70-bellemare17a}. Here the state, action value distribution is approximated by a discrete distribution with equally spaced support. Further, we employ Deep Ensembles \cite{caldeira2020deeply} to quantify the uncertainty associated with each state action pair. These ensembles are constructed using bootstrap estimates, as explained in the methods section.

\subsection*{Reinforcement Learning in Medicine} Reinforcement Learning  has been used for various healthcare applications. References \cite{yu2019reinforcement} and \cite{liu2020reinforcement} provide comprehensive surveys of healthcare and critical care applications respectively. In the specific context of sepsis treatment, Komorowski \emph{et al.} \cite{komorowski2018artificial} used a discrete state representation created by clustering patient physiological readouts, and a 25 dimensional discrete action space to compute optimal treatment strategies using dynamic programming based methods. Others have considered continuous state representations \cite{raghu2017deep} and partial observability \cite{peng2018improving}. 

Our proposed decision support system is based on a preference score as shown in Figure \ref{fig:overall}(a). In contrast to previous work, we choose a lower dimensional action space (9 actions), to ensure sufficient coverage in the training data, and a reduced decision time-scale, to be more aligned with clinical practice. The short time scale also provides a clinical justification for the less granular action space. Our rewards are based on previous work \cite{raghu2017deep} (see Methods), which has intermediate SOFA-based rewards, and $\pm 15$ terminal rewards, depending on survival.



\section{Results}
\subsection{Trajectory reconstruction using a physiology-driven autoencoder} 
\begin{figure}[ht!]
    \centering
    \includegraphics[scale=0.10]{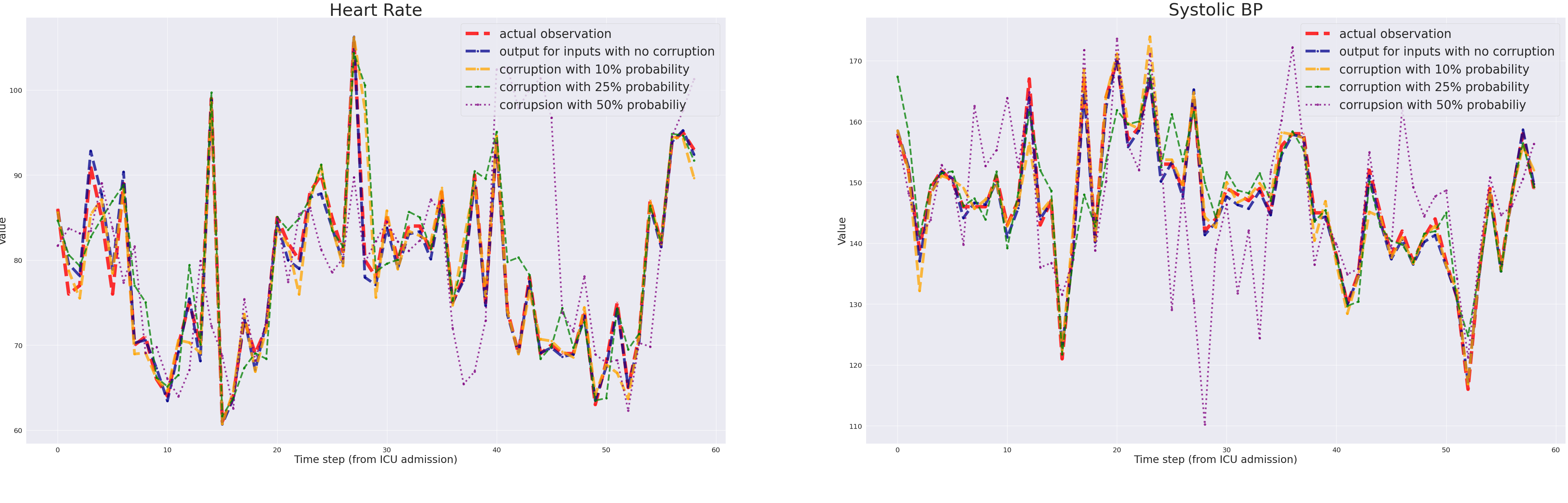}
    \caption{Reconstruction of two validation patient trajectories using different levels of corruption using the physiology-driven autoencoder, Left: Heart Rate. Right: Systolic Blood Pressure.}
    \label{fig:my_label}
\end{figure}

One of the key features of our method is the physiology-driven structure of the autoencoder that represents the cardiovascular state of the patient (see Figures \ref{fig:overall}(b) and (c)). The decoder of this autoencoder is a set of algebraic equations that map the latent state to observable, and clinically relevant physiological parameters, such as heart rate and blood pressure. Figure \ref{fig:my_label} shows selected reconstructed trajectories for one representative patient, using various levels of data corruption (see Methods). As the figure illustrates, the model  successfully reconstructs the observable outputs and their trends with corruption probabilities as high as 25\%. It is only at extreme levels of corruption (50\%) that the model's accuracy degrades. Such robustness to moderate levels of corruption was typical among training and validation patient trajectories. We thus conclude that the autoencoder has learned an effective representation of the cardiovascular state of the patient.

\subsection{Value distributions and expected values}
\begin{figure}[htp!]
    \centering
    \includegraphics[scale=0.20]{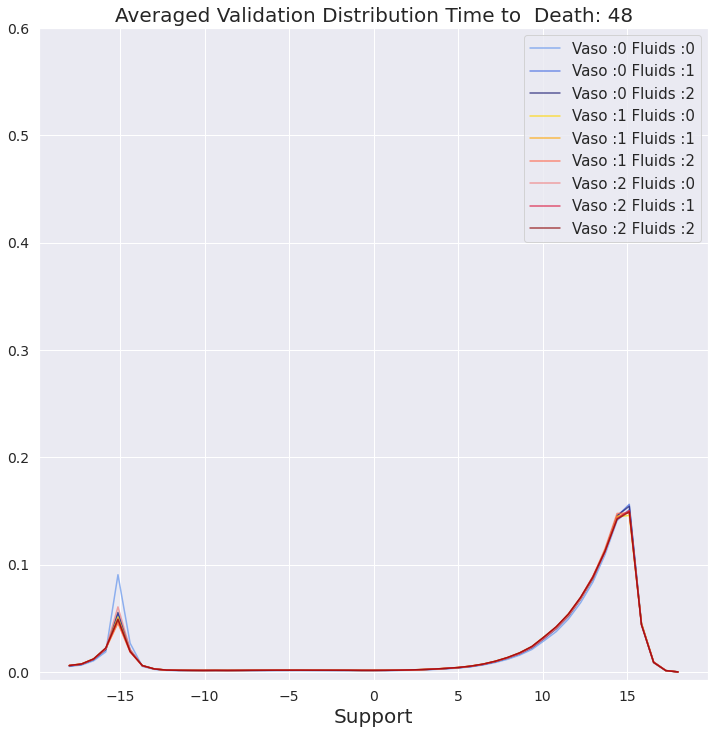}
    \includegraphics[scale=0.20]{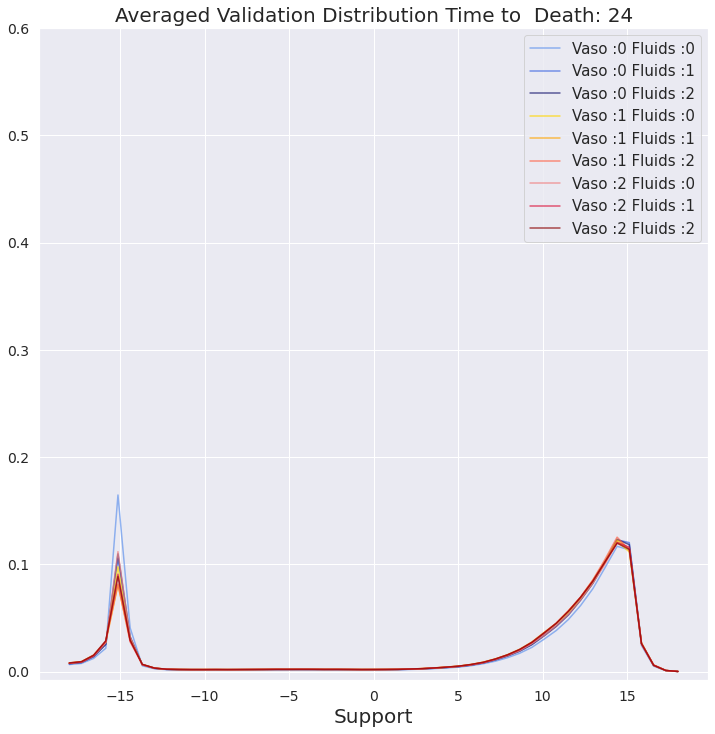}
    \includegraphics[scale=0.20]{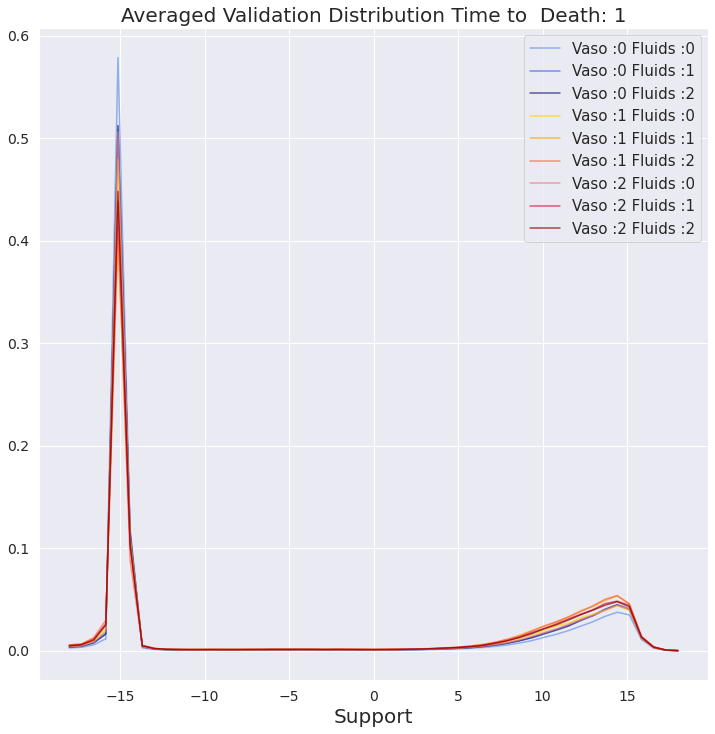}

    
    \centering
    \includegraphics[scale=0.20]{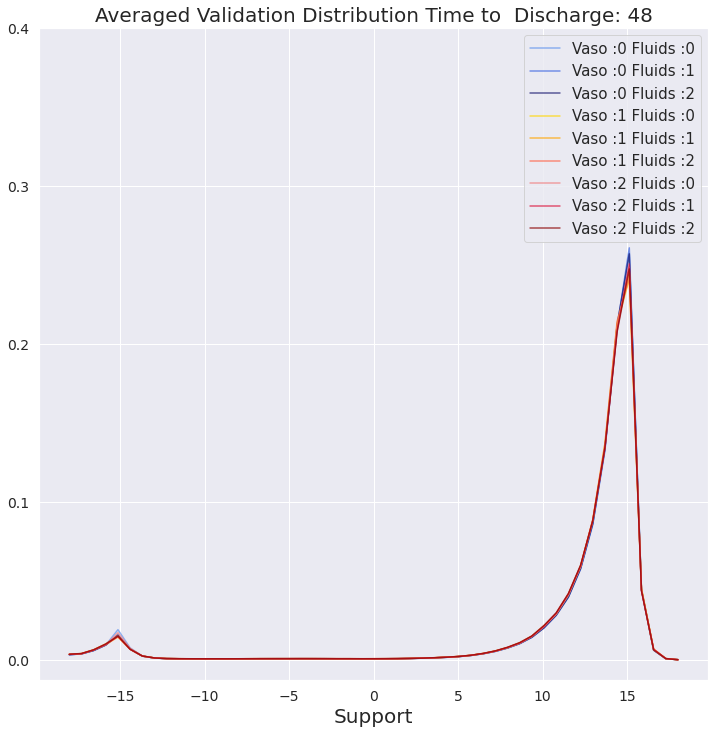}
    \includegraphics[scale=0.20]{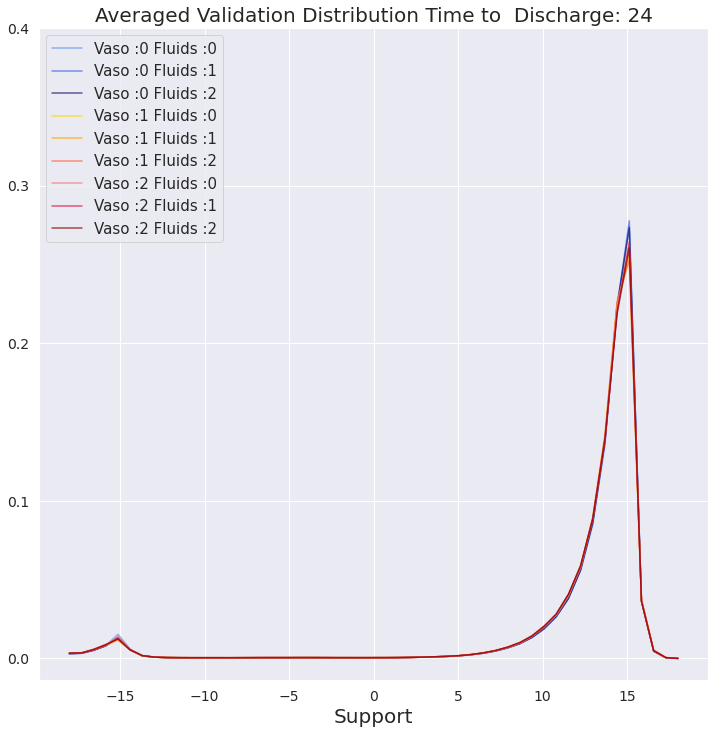}
    \includegraphics[scale=0.20]{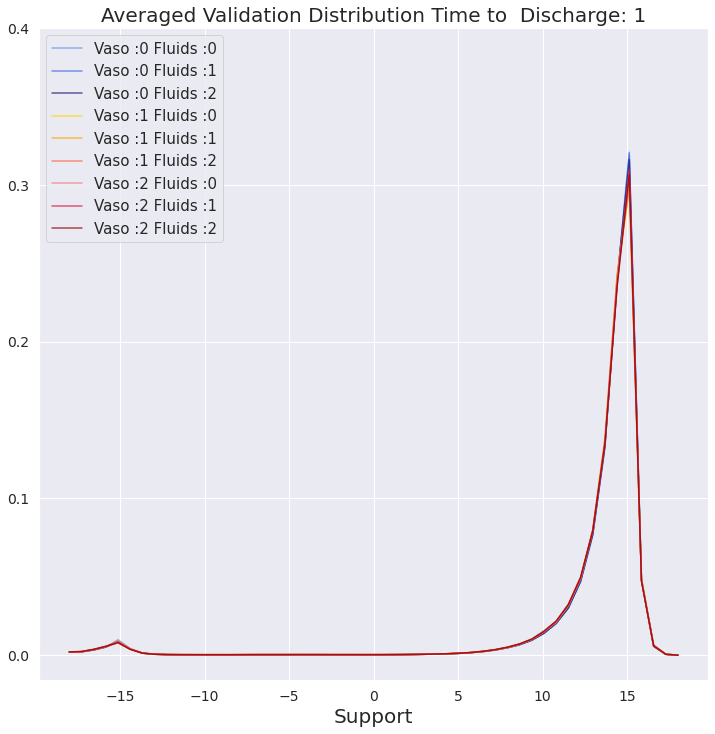}
    
    \caption{Value distributions for validation patients averaged according to  different times from death or discharge, \textbf{Top}: Non Survivors. \textbf{Bottom}: Survivors.}\footnote{In each plot, the y axis is adjusted to match the highest peak, the axes are not uniform, to better present the difference between action distributions}
    \label{fig:val_dists}
\end{figure}

We next investigated whether the learned values are generalizable, consistent with clinical knowledge, and correlated with the risk of death in non-survivors. To do this, we examined the value distributions that are produced at each time-step for patients in the validation set, stratified by outcome (i.e., survivor vs non-survivor). Figure \ref{fig:val_dists} plots the average value distributions output for non-survivors (top) and survivors (bottom) at 48, 24, and 1 hour from death or discharge. The individual lines in each panel correspond to the value distributions under the nine discrete actions available to the agent. We emphasize that these plots were generated for the purpose of analyzing the learned models. In particular, the network only sees the current state when it outputs such distributions; it is not given with any information about the future.

\begin{figure}[ht]

\includegraphics[scale=0.26]{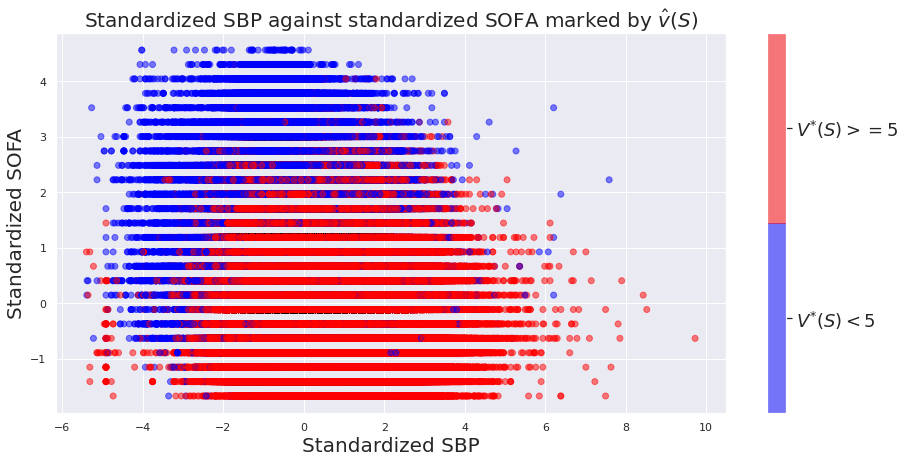}
\hspace{10 pt}
\includegraphics[scale=0.26]{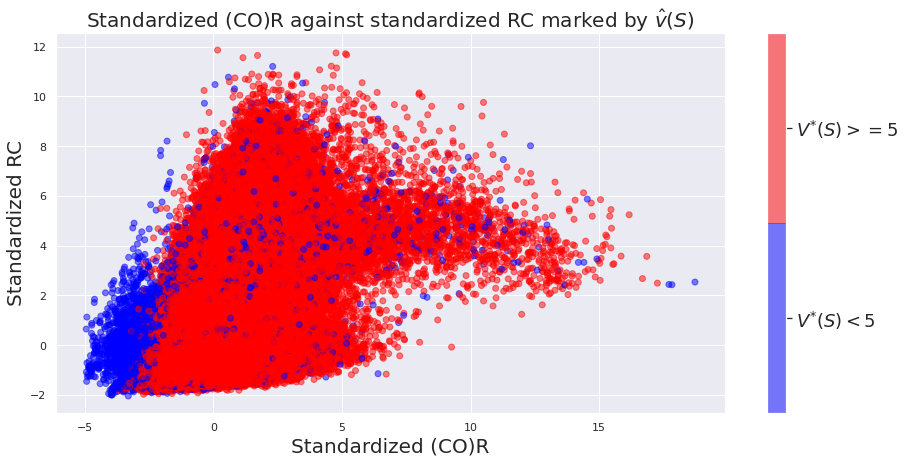}



\vspace{25 pt}

\begin{center}

\centering
 \begin{tabular}{||c | c ||} 
 \hline
 Feature & Importance Score \\ [0.5ex] 
 \hline\hline
 CNS  & 0.664190\\
BUN  & 0.462304\\
SOFA &  0.418309\\
HR    &   0.362906\\
HB     &   0.264061\\
l\_3     &  0.253673\\
R       &  0.226833\\
SV      &  0.224609\\
MBP     &  0.224562\\
AG      &  0.214475\\ [1ex] 
 \hline
 \end{tabular}

\vspace{10 pt}
\end{center}

\caption{Scatter plots of scaled features : \textbf{Top}: Marker colors indicates if $\hat{V}^{*}(S)< 5$ (Blue) or $\hat{V}^{*}(S)\ge 5$ (Red)  \textbf{Bottom}: Table 1: Top 10  features measured by feature permutation. Here, $l\_k$ denotes the kth component of the latent lab representation.}

\label{fig:feat_treat}
\end{figure}

Figure \ref{fig:val_dists} clearly exhibits bi-modal distributions over values for non-survivors as much as 48 hours in advance of death. Further, as the patient gets closer to death, the mass shifts towards the left peak (which corresponds to death). This behavior is consistent with the patient's deteriorating condition. Additionally, the distribution associated with the ``no treatment'' action has a larger left peak than others, highlighting that for these states the lack of treatment for even one hour can be fatal. The mass of the distributions for survivors, in contrast, is concentrated closer to the right limit \emph{and} there is little difference between actions. Both of these observations are consistent with the expectation that survivors are less likely than non-survivors to enter the highest risk states, and so the consequences of a change in action/treatment are less extreme.

We then investigated the dependence of features and inferred states on the value distributions and determined that they are explainable, and consistent with clinical expectation. For example, Figure \ref{fig:feat_treat} shows two scatter plots contrasting representative pairs of variables, stratified by an optimal expected value threshold of five\footnote{This threshold was chosen arbitrarily, and we could observe similar results for any reasonable threshold.}. It is clear that the model associates different states with different expected rewards/risk. For example, the model associates low SBP (hypotension) and high SOFA scores with an increased risk of death, which is consistent with medical knowledge. Thus the agent has learned to discriminate between low and high risk states in an explainable manner. The ability to learn such associations is noteworthy because the training and test data are highly imbalanced. In particular, 89\% of states have the property $V^{*}\ge 5$. 


Finally, we quantified the importance of each feature using feature permutation \cite{molnar2020interpretable}. Briefly, for each patient we permute a selected feature while keeping others fixed. The mean absolute value difference of the $Q$ function (across states and actions) is taken as the importance score for that patient. Table 1 lists the top 5 features across the entire cohort. The complete feature ranking can be found in the supplementary materials. The cardiovascular states and the latent lab representations are among the most important features, highlighting the importance of representation learning.  

\subsection{Vasopressor Treatment Strategies}

\begin{figure}[t!]
\includegraphics[scale=0.30]{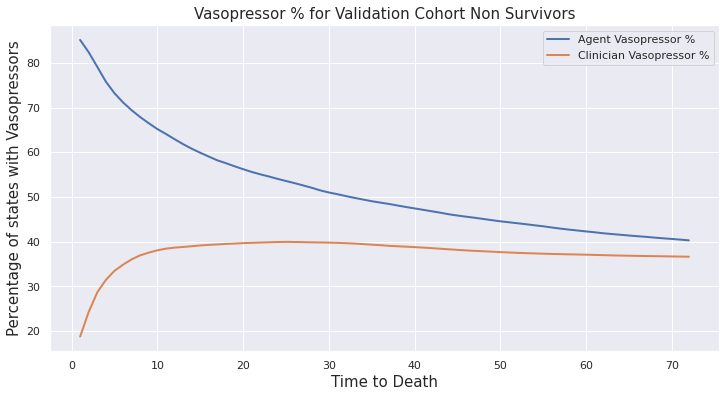}{a}
\includegraphics[scale=0.30]{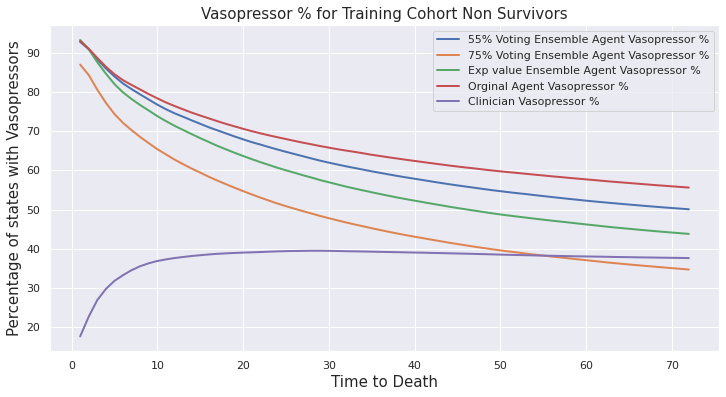}{b}

\vspace{15pt}

\includegraphics[scale=0.30]{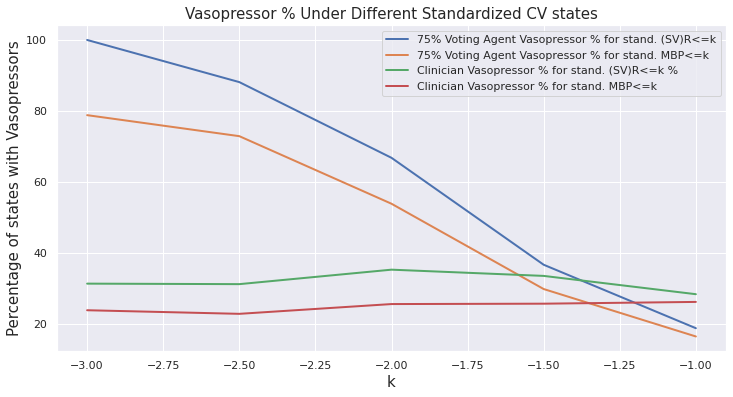}{c}
\includegraphics[scale=0.30]{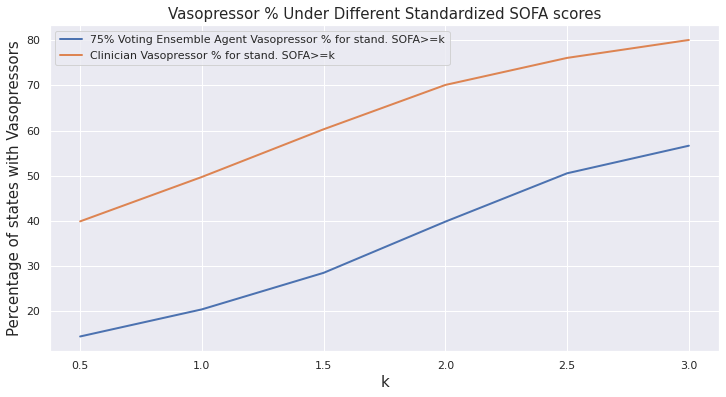}{d}

\caption{\textbf{Top}: Percentage of states with vasopressors recommended for the training and validation states, with time to eventual death. Here a $p$\% voting agent, denotes an agent which only prescribes vasopressors if an only if least $p$\% of the Bootstrapped Ensembles have agree on giving vasopressors. 
\textbf{Bottom}: The percentages of states with vasopressors recommended or given with respect to cardiovascular states and SOFA score.
}\label{fig:vaso}

\end{figure}

We observed that the RL agents consistently recommend vasopressors for near-death (non-survivor) states, and that the percentage of such states increase closer to the patient's eventual death. This phenomena is also shared by validation cohort states, as illustrated in Figure \ref{fig:vaso}(a), suggesting that this behavior isn't due to overfitting. In contrast, clinicians have only administered vasopressors on average around 40\% of the time, and this number drops off  rapidly in the last 10 hours. We investigated whether these differences are an artifact of our choice of method by evaluating different training options and algorithms. Specifically, we: (i) trained networks with and without weighted experience sampling scheme (explained under Methods); (ii) used a different distributional RL algorithm, called Quartile Regression Q Learning \cite{dabney2018distributional}; (iii) considered an artificial voting ensemble agent, which only administers vasopressors if at least $p$\% of the ensemble agrees on giving vasopressors, at a given state; and (iv) consider the expected value of the ensemble agent, which takes a weighted average (weighted by the number of patients it's trained on) of expected values of each bootstrapped network. In each case we observed similar results, as shown in Figure \ref{fig:vaso}(b).

We also investigated the relationship between vasopressor recommendation and cardiovascular states, and SOFA score. As illustrated in Figure \ref{fig:vaso}(c), the RL agents recommend vasopressors, much more regularly as (SV)R (product of stroke volume and resistance) and mean blood pressure drop. This is consistent with physiological knowledge, and latest critical care research. For example, \cite{foulon2018hemodynamic} shows that hemodynamic effects of norepinephrine extends beyond blood pressure, and it effects SV and CO, and as described earlier, increasing systemic vascular resistance and blood pressure, are among the primary goals of vasopressor therapy. However, it is interesting to note that the clinicians have not necessarily associated lower blood pressure, or (SV)R with more frequent vasopressor administration. However they do seem to give vasopressors more regularly as SOFA score increase. These results could potentially provide an important direction and hints towards \emph{better} treatment strategies.

This difference between the AI agent and human physicians is not unexpected, and does not imply that physicians are systematically acting sub-optimally. Rather, this difference reflects the fact that the rewards that the agents were trained on only consider the final state of the patient.  They do not, for example, incorporate decisions that were made by the patient's family to cease extraordinary measures, after consultation with the physician. Such status changes are common, but were not available in the training data. 


In contrast to vasopressors, RL agents and clinicians had similar frequencies of fluid administration for non-survivors. However, there were some disagreement even amongst the ensembles on whether or not to administer fluids for survivors (at less risky states). We present a more detailed analysis along with global results in the supplementary information.

\begin{figure}[t]
\centering
\includegraphics[width=220pt]{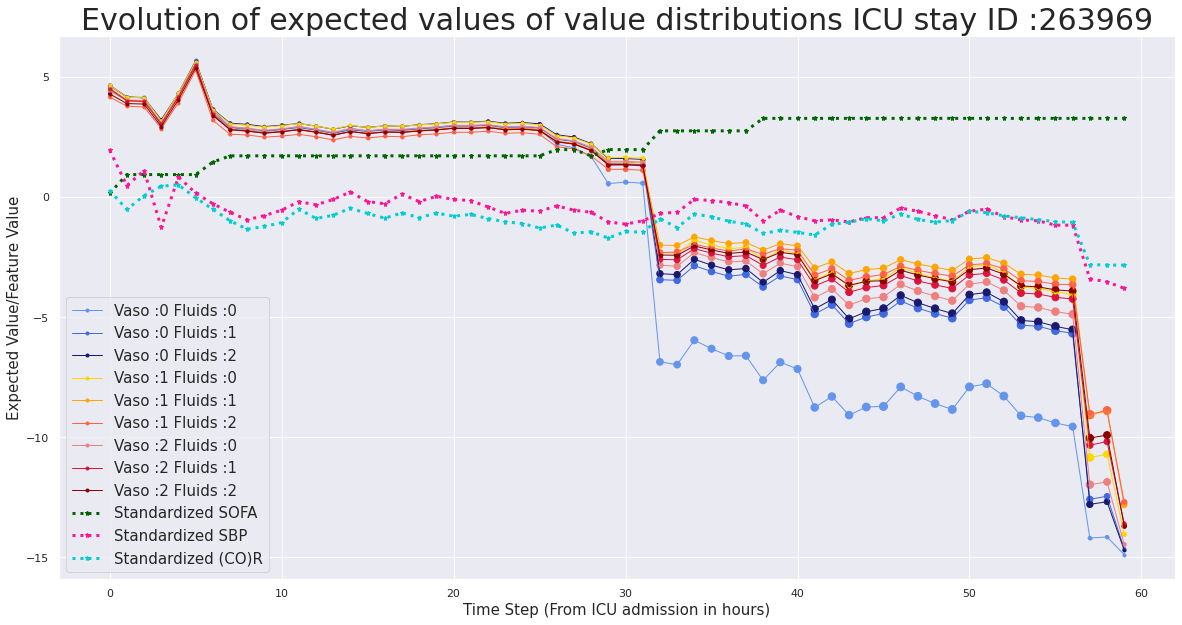}{a}
\includegraphics[width=220pt]{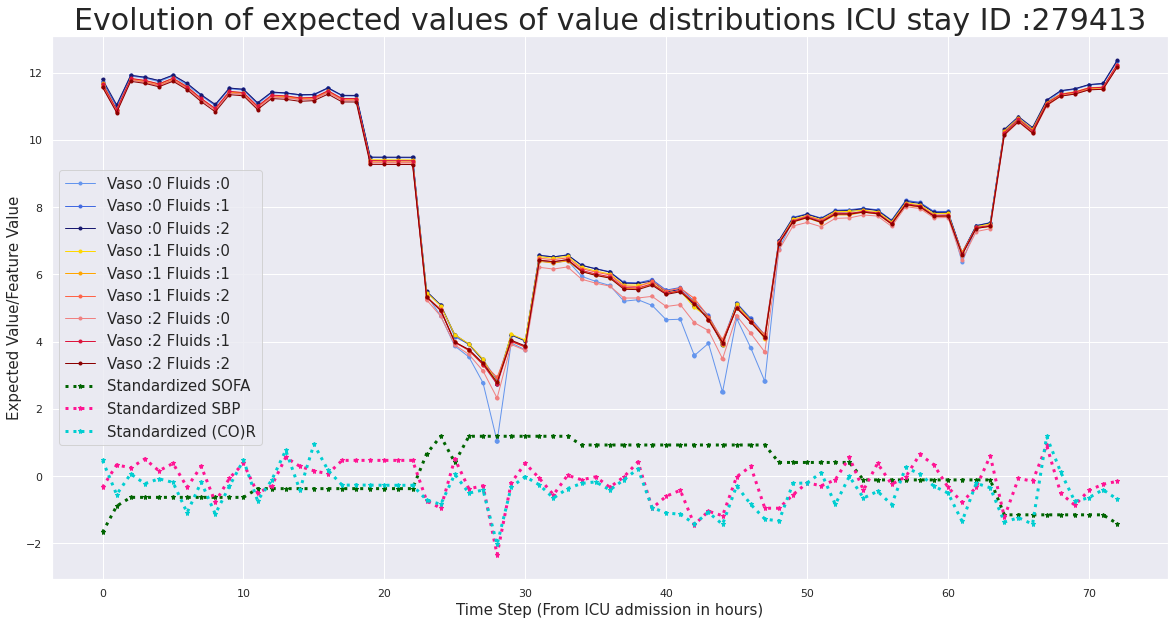}{b}


\centering

\includegraphics[scale=0.18]{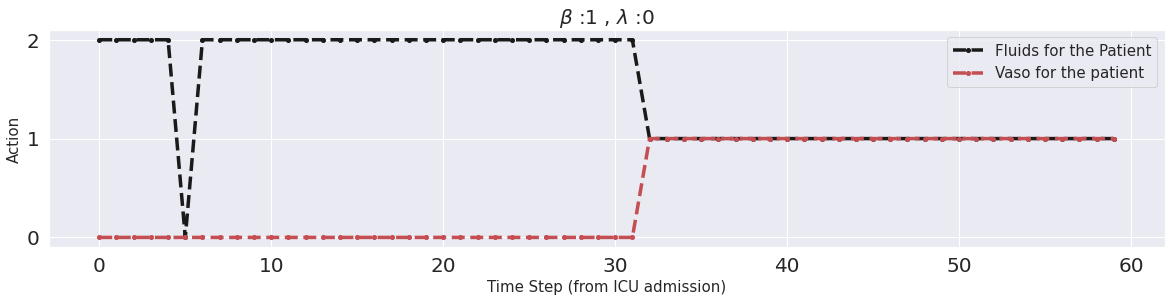}{c}
\includegraphics[scale=0.18]{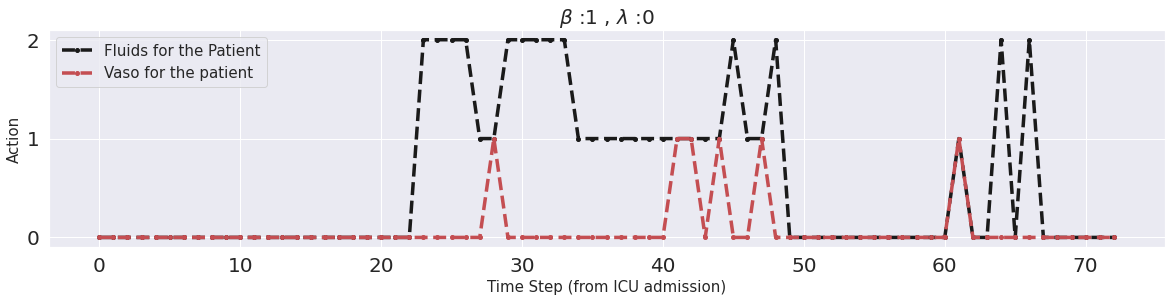}{d}

\includegraphics[scale=0.18]{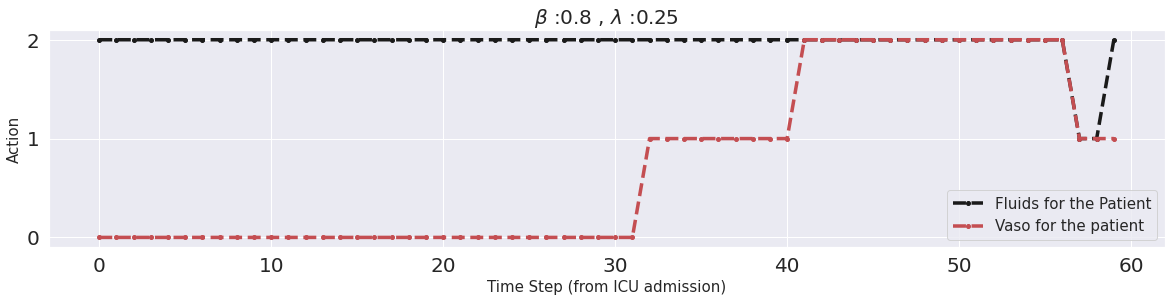}{e}
\includegraphics[scale=0.18]{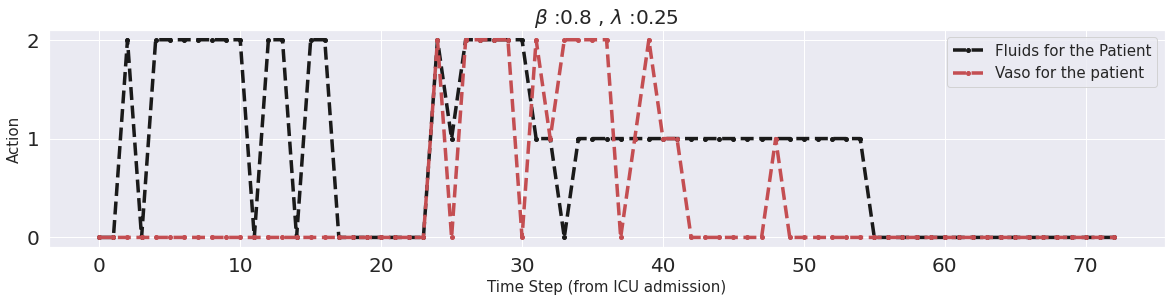}{f}

\includegraphics[scale=0.18]{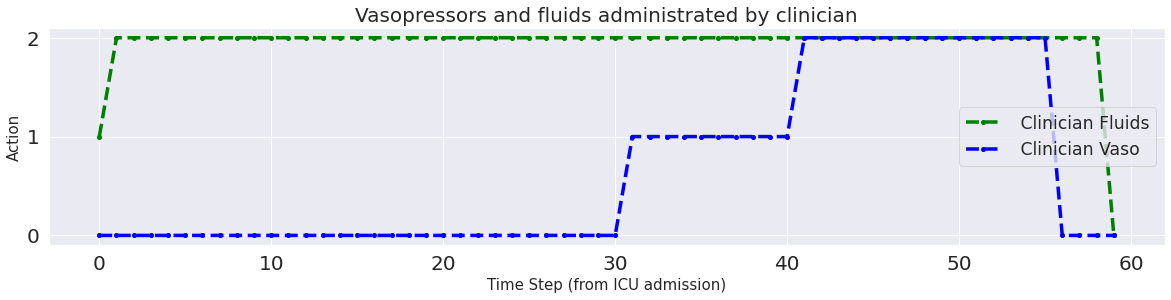}{g}
\includegraphics[scale=0.18]{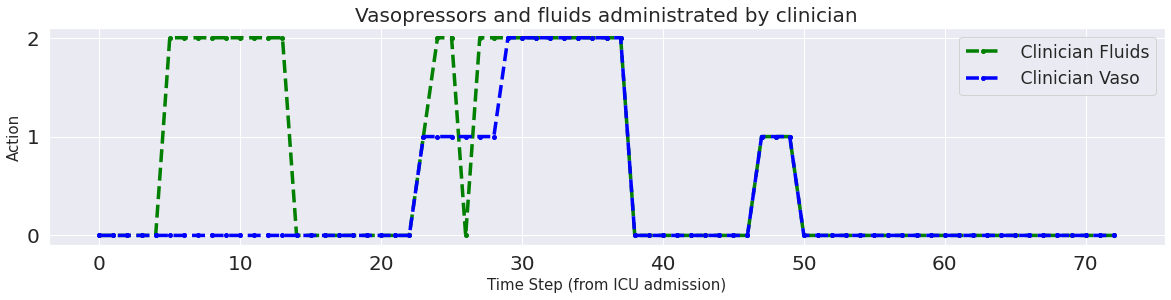}{h}





\hspace{25pt}

\caption{\textbf{Expected value evolution of the main agent for two patients} \textbf{(a)}: A patient who died in the ICU.  The marker size indicates the parametric uncertainty associated with a particular action. Also shown are the standardized values of SOFA score, Systolic blood pressure, and the unidentifiable cardiovascular state (CO)R. The $x$-axis indicates the hours from ICU admission. 
\textbf{(b)}: A survivor.  \textbf{Recommended treatments under various preference parameters} (see Eq. \ref{eq:pref}). \textbf{(c,e)}: Recommendations for the same patient as (a). \textbf{(d,f)} Recommendations for the same patient as (b).\textbf{Actual clinician treatments (g,h)}: for the patients in (a) and (b).}

\label{fig:exp_dead}

\end{figure}
\subsection{Uncertainty aware treatment}

Next, we consider representative patients, and analyze the expected values of all distributions and model uncertainty. Figure \ref{fig:exp_dead}(a) shows the evolution of expected values for a non-survivor (ICU ID: 263969). This was typical among all non-survivors; initially there's less variability among the expected values, but as the patient's health deteriorates the variation becomes more drastic, and there is a clear preference towards vasopressor-based actions. The marker size indicates how much the agent is uncertain of its own results. We observe that the model is less certain when the patient's health starts deteriorating. This can be attributed to the fact that these states are uncommon in the training data, and that the underlying cause driving deterioration can vary widely in septic patients.

For comparison, Figure \ref{fig:exp_dead}(b) shows the expected values of a survivor (ICU ID: 279413). Here the expected values take a downward slide at around 25 hours from admission, with the values associated with no treatment considerably lower. This coincides with SOFA score increasing and SBP (CO)R rapidly decreasing, clearly indicating that the patient's health is deteriorating. However, as SOFA score improves and the pressure and (CO)R goes up, the expected values do go up, and the difference between expected values of each distribution is considerably less. The uncertainty levels are also much lower.

The fact that expected values of different actions are close to each other in \textit{healthy} patient states can be explained by equation \ref{eq:bell_2}. State-action values are calculated under the assumption that the agent always takes the optimal action.  Our agent chooses an action every hour, and the intermediate rewards are much smaller in value than the terminal rewards. Thus, the value of the choice of action is not likely to change very much in a \textit{healthy} patient state from hour to hour. Put another way, any mistake made by the agent is easily reversed by taking the correct action in the next hour if the patient is non-critical. In contrast, in more critical states, a wrong action can have irreversible consequences. 

Figures \ref{fig:exp_dead}(c)-(f) show different treatment recommendations under our proposed framework for uncertainty-aware decision support. Briefly, the user specifies their relative confidence in the RL-agent and a behavior cloner (which represents the human agent) by specifying a parameter, $\beta$. Lower values of $\beta$ place more emphasis on the behavior cloner. An action preference score (see. Methods, Eq. \ref{eq:pref}) is then calculated for each action in the current state. The score is a simple mixture of the (scaled \footnote{We do this using a softmax function}) expected value of the ensembled distribution and the behavior probability, discounted by the model uncertainty corresponding to the state-action pair, using a parameter $\lambda$. Panels (c)-(f) illustrate that different choices are made, depending on the value of $\beta$ and $\lambda$. Further, the sequence of treatments are qualitatively different for the non-survivor (panels c and e) and the survivor (panels d and f), because the agent has learned to identify critical states that require interventions; the average non-survivor tends to remain in such states for longer stretches, and so the agent makes relatively few adjustments, compared to the survivor. Once again, the agent does not know the ultimate fate of the patient. For comparison, panels (g) and (h) show the actual clinician treatments for the two patients.  We present more detailed analyses of the RL treatments in the Supplemental Material.

\subsection{A comment on Off Policy Evaluation}
Off policy evaluation (OPE) is the quantitative or statistical evaluation of the value of a learned policy, usually using another dataset. Although attractive in theory, most unbiased OPE methods use importance sampling, and are therefore dependent on a known behavior policy. This is not the case when the data were generated by human clinicians. Even if a suitable behavior policy \emph{were} known, an obviously bad policy can result in a very high OPE value in our setting. For example, an agent that always prescribes no treatment for critical patients would, in effect, eliminate most of the rewards accumulated by non-survivors which are, of course, the source of the majority of the \emph{negative} rewards. Such a policy would have a misleadingly high OPE, because human clinicians rarely withhold treatment for critical patients (the one exception being a conscious decision by the family to terminate extraordinary interventions), and so the the importance weights for such trajectories will tend towards 0.


We note that previous research has also argued at \emph{all} OPE methods are unreliable in the context of sepsis management, and state-of-the-art OPE methods may fail to differentiate between obviously good and obviously bad policies \cite{gottesman2018evaluating}. Thus, we do not consider statistical OPE in this work.  However, we note that developing OPE techniques suited for the critical care domain is an important area of research to explore in the future.


\section{Discussion and Conclusion}
We present an interdisciplinary approach which we believe takes a significant step towards improving the current state of data-driven interventions in the context of clinical sepsis, in terms of improving both outcome and interpretability. Indeed, we believe that the maximum benefit of Artificial Intelligence applied to medicine is best realized through the integration of mechanistic models of physiology whenever possible, uncertainty quantification, and human expert knowledge into sequential decision making frameworks.

Our contribution improves the status quo in several ways. Compared to prior work, our approach deals with partial observability of data, yet known physiology, by leveraging a low-order two-compartment Windkessel-type cardiovascular model in the context of self-supervised representation learning. As mentioned previously, this has several benefits. First, in the context of sepsis treatment, estimating the cardiovascular state is essential because the clinical decision to administer intravenous fluids or vasopressor is driven by an implicit differential diagnosis by the clinician, as to whether insufficient organ perfusion and shock are secondary to insufficient circulating volume (thus requiring fluids), vasoplegia (thus requiring vasopressors), or some combination of both fundamental pathophysiologies. Second, there is typically insufficient data to determine whether heart function is adequate (contractile dysfunction), but a mechanistic model provides an indirect means for estimating cardiac function by imposing known physiology. Finally, the incorporation of physiologic models improves model explainability, while deep neural networks and stochastic gradient-based optimizers make it possible to learn robust and generalizable representations from large data. We expect the unification of models based on first-principles and data-driven approaches will provide a powerful interface between traditional computational sciences and modern machine learning research, mutually benefiting both disciplines.

We also introduce an approach to quantifying model uncertainty, which is essential in any practical application of RL-based inference using clinical data. To the best of our knowledge, this is the first time uncertainty quantification is used to quantify epistemic uncertainty in RL-based optimization of sepsis treatment, and of critical care applications more generally. \footnote{Previous approaches ex. \cite{li2019optimizing} have considered inherent environment uncertainty}. The method's uncertainty estimates, combined with the recommended action comprise a simple framework for automated clinical decision support. This principle  aligns with the larger goal of combining different forms of expertise and knowledge for better decision making, a philosophy consistent with the rest of this work.

We chose a decision time step of one hour. Compared to similar work, this is much more compatible with the time scale of medical decision making in sepsis, where fluid and vasopressor treatments are titrated continuously. Accordingly, on such a time scale, there does not appear to be large differences in the relative merit of different dosing strategies. This makes intuitive sense: there is presumably a lesser need for major treatment modifications if decisions are made more frequently. Yet, a frequent finding across patients, especially the sickest ones, was that inaction (no intervention) was a consistently worse strategy. This also meets clinical intuition.

Reducing the time scale of decisions is not only appealing clinically in situation of rapidly evolving physiological states, such as is the case in early sepsis, but it also provides a more compelling basis for a less granular action space. Indeed, if decisions are made hourly, it does meet clinical intuition to have fever discrete actions. Few physicians will argue that there is likely to be little difference in administering 100cc or 200cc of fluids in the next hour. In the extreme, if time were continuous, the likely decision space at any given time, is whether a fluid bolus should be administered or not. A similar reasoning applies to vasopressors (increase, reduce, status quo). We further notice that our methods consistently identify high risk, non-survivor patient states which can \emph{potentially} benefit from more frequent vasopressor treatment. These results should of course, be subject to clinical verification. 

An important open problem in the application of offline RL to medicine is the means by which one evaluates  learned treatment policies, given the obvious ethical issues associated with allowing an AI to exert some control over treatment. Still, proper clinical trials will be necessary, eventually, so the critical care community should define for itself the standards by which an AI would be deemed safe enough to enter clinical trials \cite{rivera2020guidelines}. In this work, we have largely relied on a combination of medical expertise, and the fact that our model leverages prior knowledge in the form of a simple model of cardiovascular physiology, to argue that the learned policy is reasonable. We make no claim that the policy is expected to produce superior outcomes in sepsis patients, relative to human clinicians.  One important area for future work may be the incorporation of more detailed models of physiology into our framework, or perhaps using such models in the context of \emph{in silico} trials (ex. \cite{Clermont2004}) as a first step towards demonstrating that a learned policy is safe, and perhaps suitable for pre-clinical and clinical trials. Additional areas for future work include the design of alternative rewards (ex. based on time-dependent hazard ratios for death), and the application of risk-averse offline RL (ex. \cite{Urpi21}).

\section{Methods}
\subsection{Data sources and preprocessing}
Our cohort consisted of adult patients ($\ge 17)$ who satisfied the Sepsis 3 \cite{johnson2018comparative} criteria from the Multi-parameter Intelligent Monitoring in Intensive Care (MIMIC-III v1.4) database \cite{mimiciii}, \cite{mimiciiidata}. We excluded patients with more than 25\% missing values after creating hourly trajectories, and patients with no weight measurements recorded. The starting point of trajectories is ICU admission.

We further excluded patients who got discharged from the ICU but ended up dying a few days or weeks later at the hospital. Since we don't have access to their patient data after the ICU release, treating the final ICU data as a terminal state would damage generalizability. We cannot treat those patients as survivors, however, as they were not released from the hospital.

Actions were selected by considering hourly total volume of fluids (adjusted for tonicity), and norepinephrine equivalent hourly dose (mcg/kg) for vasopressors. In computing the equivalent rates of each treatment, we followed the exact same queries as Komorowski et al \cite{komorowski2018artificial}. When different fluids were administrated, we summed up the total fluid intake within the hour, and discretized the resulting distribution. For vasopressors, we considered the maximum norepinephrine equivalent rate administered within the hour to infer the hourly dose. We used 0.15 mcg/kg/min norepinephrine equivalent rate, and 500 ml for fluids, as the 1,2 cutoff when discretizing.\footnote{These were chosen, considering the mean, median of non zero rates and medical knowledge, We also observe that due to the low dimensional action space, there is flexibility in choosing the cutoffs}. A separate 0 action for each was added to denote no treatment.

Missing vitals and lab values were imputed using a last value carried forward scheme, as long as missingness remained less than 25\% of values. A detailed description on extracting, cleaning and implementation specific processing as well as additional cohort details are included in the appendices.

\subsection{Models}
\subsubsection{Physiology-driven Autoencoder}

Autoencoders are a type of neural networks which learn a useful latent, typically lower-dimensional representation of input data, while assessing the fidelity of this representation by minimizing data reconstruction error. Our autoencoder architecture provides an implicit regularization by constraining the latent states to have physiological meaning, and the decoder to be a fixed physiologic model described in the next section. We further use a denoising scheme by randomly zeroing out input with a probability of 10-25\% , when feeding into the network. This random corruption forces the network to take the whole patient trajectory (prior to the current time point) and previous treatment into account when producing its output, because it prevents the network from \textit{memorizing} the current observation. In essence, we ask the inference network to predict observable blood pressures and the heart rate using corrupted versions of itself, by first projecting it into the cardiovascular latent state, and then decoding that to reconstruct.

Figure \ref{fig:overall}(b), shows the complete architecture of our inference network. As shown in the figure, the encoder is comprised of three neural networks, a patient encoder which computes initial hidden state estimates, a gated recurrent unit (GRU) \cite{cho2014learning} based recurrent neural network to encode the past history of vitals and scores up to and including the current time point, and a transition network which takes the previous state, the action and the history representation to output new cardiovascular state estimates. We train this structure end-to-end by minimizing the reconstruction loss, using stochastic gradient-based optimization. The supplementary material provides a detailed description of model and architecture hyper-parameters, and training details.

\paragraph{Cardiovascular Model}
The cardiovascular model, is based on a two-element Windkessel model illustrated using the electrical analog in Figure \ref{fig:overall}(c). This model provides a lumped representation of the resistive and elastic properties of the entire arterial circulation using just two elements, a resistance $R$ and a capacitance $C$, which represent the systemic vascular resistance (SVR), and the elastance properties of the entire systemic circulation, respectively. Despite it's simplicity, this model has been previously used to predict hemodynamic responses to vasopressors \cite{bighamian2013analytic} and as an estimator of cardiac output and SVR \cite{bighamian2014prediction}.

The differential equation representing this model is:
   \begin{equation}
       \frac{\mathrm{d}P(t)}{\mathrm{d}t}=-\frac{1}{RC}P(t)+\frac{Q(t)}{C}
   \end{equation}

were $Q(t)$ represents the volume of blood in the arterial system. As explained in \cite{bighamian2013analytic}, over the interval $[0,T]$ (where $T$ is the filling time of the arterial system) we can write $Q(t)$ as $Q(t)=SV\delta(t)$,
where $SV$ stands for Stroke Volume, the volume of blood ejected from the heart in a heartbeat. When the system is integrated over the interval $[0,T]$ we obtain the following expressions for $P_{sys}; P_{dias}; P_{MAP}$ , i.e., the systolic,diastolic, mean arterial pressure, respectively,

 \begin{equation}
 P_{sys}=\frac{SV}{C}\frac{1}{1-e^{-T/RC}}, \hspace{10pt}
    P_{dias}=\frac{SV}{C}\frac{e^{-T/RC}}{1-e^{-T/RC}}, \hspace{10pt} P_{MAP}=\frac{(SV)R}{T}=\frac{(SV)FR}{60}
   \end{equation}

$T$ is the filling time and $F$ is the heart rate, which is determined by $T$. This system of algebraic equations is used for the decoder of our autoencoder. Since heart rate can itself be affected by vasopressors and fluids, we added heart rate ($F$) as an additional cardiovascular state despite it being observable.

Therefore we have a multivariate function $f:\{R,C,SV,F,T\} \to \{P_{sys},P_{dias},P_{MAP},F\}$, represented by the equations above, and the trivial relationship $F=F$
(Despite the obvious relationship we used both $F$ and $T$, for ease of training and stability.) As stated previously, to prevent it from just using the current observations, we use a denoising scheme for training. This ensures at a fixed time, the model cannot \textit{memorize} the current observation and learn to invert $f$, since there is a nonzero probability of corruption. Thus it has to learn to factor in the history and the treatments when determining the cardiovascular states. Once $SV$ is inferred, the cardiac output (CO), can be computed as $CO=(SV)F$.

Since $f$ is not one to one, typically not all states are identifiable. 
To arrive at a better approximation we used the latent space to only model deviations from fixed baselines. We also posit that identifiable combinations of states, when trained with a denoising scheme, should provide important cardiovascular representations in the POMDP setting.

\subsubsection{Denoising GRU autoencoder for representing Lab history}

We use another recurrent autoencoder to represent patient lab history, motivated by the fact that labs are recorded only once every 12 hours. Forward filling the same observation for 12 time points, is almost certainly sub-optimal, and the patterns of change in lab history can be helpful in learning a more faithful representation. Thus, we use a denoising GRU autoencoder constructed by stacking three multi-layer GRU networks on top of each other, with a decreasing number of nodes in each layer, the last 10 dimensional hidden layer was used as our representation. This architecture is motivated by architectures used in speech recognition \cite{chan2016listen}.

This model was also trained by corrupting the input, where each data-point was zeroed with a probability of up to 50\%. (The rate was gradually increased from 0 to 50\%). As with the previous autoencoder, this provides an extra form of regularization, and forces the learned representation to encode the entire history.

Model architecture and training details and presented in the supplementary materials.

\subsubsection{Behavior Cloner}
We use a standard multi-layer neural network as our imitation learner. This model is trained using stochastic gradient-based optimization by minimizing the negative log-likelihood loss, between the predicted action and the observed clinician action, with added regularization to prevent overfitting. 

We do mention that there are many other options that could be used as a imitation learner, including nearest neighbor-based method as in \cite{peng2018improving}.

\subsection{POMDP Formulation}
\paragraph{States:} A state is represented by 41 dimensional real-valued vector consisting of:

\begin{itemize}
    \item \textbf{Demographics}: Age, Gender, Weight.
    \item \textbf{Vitals}: Heart Rate, Systolic Blood Pressure, Diastolic Blood Pressure, Mean Arterial Blood Pressure, Temperature, SpO2, Respiratory Rate.
    \item \textbf{Scores:} 24 hour based scores of, SOFA, Liver, Renal, CNS, Cardiovascular
    \item \textbf{Labs:} Anion Gap, Bicarbonate, Creatinine, Chloride, Glucose, Hematocrit, Hemoglobin, Platelet, Potassium, Sodium, BUN, WBC.
    \item \textbf{Latent States:} Cardiovascular states and 10 dimensional lab history representation.

\end{itemize}

\paragraph{Actions:} To ensure each action has a considerable representation in the dataset, we discretize vasopressor and fluid administrations into 3 bins, instead of 5 as in previous work \cite{raghu2017deep}, \cite{komorowski2018artificial} \cite{peng2018improving}.
This results in 9 dimensional action space.

\paragraph{Timestep :} 1 hour

\paragraph{Rewards :}We use the reward structure that was suggested by Raghu et. al \cite{raghu2017deep}, with a minor modification. Since lactate was very sparse amongst out cohort we only considered SOFA based intermediate rewards. 
Specifically,  whenever $s_{t+1}$ is not terminal, we use reward of the form:

\begin{equation}
    r(s_{t},a,s_{t+1})=-0.025\mathbb{I}((s_{t+1}^{SOFA}=s_{t}^{SOFA} \hspace{2pt } \& \hspace{2pt }s_{t+1}^{SOFA}>0)-0.125\mathbb{I}(s_{t+1}^{SOFA}-s_{t}^{SOFA})
\end{equation},

For terminal rewards we put $r(s_{t},a,s_{t+1})=15$ for survival and $r(s_{t},a,s_{t+1})=-15$ for non-survival.

\subsubsection{Training}
We only mention important details of training the RL algorithms here. Representation Learning related training and implementations are detailed out in the supplementary information.

We train the Q networks using a weighted random sampling-based experience replay, analogous to the prioritized experienced replay \cite{schaul2015prioritized}, which has resulted in superior performance in classical DRL domains, such as Atari games.

In particular for each batch, we sample our transitions from a multinomial distribution, with higher weights given to terminal death states, near death states (measured by time of eventual death), and terminal surviving states. We used a batch size of 100, and adjusted weights such that on average there is 1 surviving state, and 1 death state in each batch.

This does introduce bias, with respect to the existing transition dataset, however we argue that this would correspond to sampling transitions from a different data distribution, which is closer to the true patient transition distribution, we are interested in, as we are necessarily interested in reducing mortality. We empirically observe that, when using such a weighting scheme the value distributions align more closely to clinical knowledge in identifying risky states, and \textit{near} death states.

A same weighting scheme was used for all ensemble networks, which are trained to estimate uncertainty. As mentioned previously, we verify that the main results on vasopressor treatment strategies hold even for pure random sampling. 

\subsection{Uncertainty}
In this section, we consider model uncertainty, and not the inherent environment uncertainty. Model uncertainty stems from the data used in training, neural network architectures, training algorithms, and the training process itself.

Inspired by statistical learning theory \cite{vapnik1992principles}, and the associated structured risk minimization problem \cite{Zhang2010StructuralRM}, we define the model uncertainty, (conditioned on a state $s$ and a action $a$), given our learning algorithm, and model architecture as :

\begin{equation}
   \mathbb{E}_{\theta,D}[l(\theta,\mathbb{E}_{D}[\theta])|s,a]
=\int l(\theta,\mathbb{E}_{D}[\theta]) \rvert_{s,a}p(\theta,D) d\theta dD =\int l(\theta,\mathbb{E}_{D}[\theta]) \rvert_{s,a}p(\theta|D)p(D) d\theta dD
\label{uncert}
\end{equation}

Here, $D$ denotes the unknown distribution of ICU patient transitions that we are attempting to learn our policies with respect to. $\theta$ is a random variable which characterizes the value distributions. (For the C51 algorithm this can be interpreted as an element in $\mathbb{R}^{51}$). This is outputted by our networks trained on a dataset sampled from $D$, for a given state action pair. This random variable is certainly dependent on the training data, and the randomness stems from the inherent randomness of stochastic gradient based optimization and random weights initialization. The quantity $l$ is a divergence metric appropriate for comparing probability distributions. We use the Kullback–Leibler divergence \cite{kullback1951information} for $l$.

\subsubsection{Estimating the uncertainty measure}
We construct a Monte-Carlo estimate of the integral in \eqref{uncert} by bootstrapping 25 different datasets, each substantially smaller than the full training dataset, and training identical distributional RL algorithms in each. This can be done efficiently due to the sample efficiency of distributional methods. Additionally, we can approximate $\mathbb{E}[\theta]$ either by the ensemble value distribution, or by the value distribution of the model trained on the full training dataset.

\subsection{Uncertainty aware treatment}
In this section, we describe a general framework for choosing actions that factors in uncertainty.  Notice that, because our RL algorithm learns (an approximation of) the optimal value distributions, making decisions by considering additional information does not violate any assumption underlying the learning process.

When suggesting safe treatment strategies, we want the proposed action to have high expected value, however we  would also like our agent to flexible enough to propose an action with less model uncertainty, if two actions have very close expected values to each other. Another important factor to consider is how likely an action is to be taken by a human clinician. This will have significance in a situation where human expertise is scarce. Large retrospective datasets subsume experience of hundreds of clinicians, and knowing what previous clinicians have done in similar situations, will be valuable such situations. Therefore we use behavior cloning to learn an approximate behavior policy of clinicians on average.


To satisfy all three goals, we propose a general framework for choosing actions, based on an action preference score, $\mathcal{P}(s,a)$, parameterized by two parameters. This general framework is flexible, yet simple, and the end-user can choose the parameters to reflect their own expert knowledge, and confidence of the framework.

Let $G(s,a)$ be a human behavior likelihood score function. In this work we equate $G(s,a)$ with the probabilities outputted by the behavior cloning network described in section 6.2.3.
Given a state $s$, we define  $\mathcal{P}(s,a)$ associated with each action $a$, as:

\begin{equation}
  \mathcal{P}(s,a)= \beta(Softmax(\tilde{Q^{*}}(s,a))+(1-\beta)G(s,a)-\lambda u(s,a)  
    \label{eq:pref}
\end{equation}

where $\beta,\lambda \ge 0$, $u(s,a)$ is the parametric uncertainty associated with the state-action pair, $s,a$, $G(s,a)$ is the behavior likelihood probability and $\tilde{Q^{*}}(s,a)$ is the $Q$ function computed from the ensembled value distributions.
When human expertise is available, $G(s,a)$ can be modified or even re-defined to factor in expert opinion. $\lambda$ penalizes uncertainty, and a low $\beta$ forces the action to be close to a clinician action. We could recover the expected value criteria by setting $\beta=1,\lambda=0$, and we could use the system as a pure behavior cloner, by setting $\beta=0,\lambda=0$. Therefore $\beta$ controls how far from the highest expected value/behavior likelihood score can the agent choose an action.

\bibliographystyle{unsrt}

\bibliography{references}{}

\begin{thebibliography}{10}

\bibitem{liu2014hospital}
Vincent Liu, Gabriel~J Escobar, John~D Greene, Jay Soule, Alan Whippy, Derek~C
  Angus, and Theodore~J Iwashyna.
\newblock Hospital deaths in patients with sepsis from 2 independent cohorts.
\newblock {\em Jama}, 312(1):90--92, 2014.

\bibitem{rhee2017incidence}
Chanu Rhee, Raymund Dantes, Lauren Epstein, David~J Murphy, Christopher~W
  Seymour, Theodore~J Iwashyna, Sameer~S Kadri, Derek~C Angus, Robert~L Danner,
  Anthony~E Fiore, et~al.
\newblock Incidence and trends of sepsis in us hospitals using clinical vs
  claims data, 2009-2014.
\newblock {\em Jama}, 318(13):1241--1249, 2017.

\bibitem{paoli_reynolds_sinha_gitlin_crouser_2018}
Carly~J. Paoli, Mark~A. Reynolds, Meenal Sinha, Matthew Gitlin, and Elliott
  Crouser.
\newblock Epidemiology and costs of sepsis in the united states—an analysis
  based on timing of diagnosis and severity level*.
\newblock {\em Critical Care Medicine}, 46(12):1889–1897, 2018.

\bibitem{marik2015demise}
PE~Marik.
\newblock The demise of early goal-directed therapy for severe sepsis and
  septic shock.
\newblock {\em Acta Anaesthesiologica Scandinavica}, 59(5):561--567, 2015.

\bibitem{lazuar2019precision}
Alexandra Laz{\u{a}}r, Anca~Meda Georgescu, Alexander Vitin, and Leonard
  Azamfirei.
\newblock Precision medicine and its role in the treatment of sepsis: a
  personalised view.
\newblock {\em The Journal of Critical Care Medicine}, 5(3):90--96, 2019.

\bibitem{douglas2020fluid}
Ivor~S Douglas, Philip~M Alapat, Keith~A Corl, Matthew~C Exline, Lui~G Forni,
  Andre~L Holder, David~A Kaufman, Akram Khan, Mitchell~M Levy, Gregory~S
  Martin, et~al.
\newblock Fluid response evaluation in sepsis hypotension and shock: A
  randomized clinical trial.
\newblock {\em Chest}, 2020.

\bibitem{sutton1998rli}
Richard~S. Sutton and Andrew~G. Barto.
\newblock {\em Reinforcement Learning: An Introduction}.
\newblock MIT Press, 1998.

\bibitem{komorowski2018artificial}
Matthieu Komorowski, Leo~A Celi, Omar Badawi, Anthony~C Gordon, and A~Aldo
  Faisal.
\newblock The artificial intelligence clinician learns optimal treatment
  strategies for sepsis in intensive care.
\newblock {\em Nature medicine}, 24(11):1716--1720, 2018.

\bibitem{raghu2017deep}
Aniruddh Raghu, Matthieu Komorowski, Imran Ahmed, Leo Celi, Peter Szolovits,
  and Marzyeh Ghassemi.
\newblock Deep reinforcement learning for sepsis treatment.
\newblock {\em arXiv preprint arXiv:1711.09602}, 2017.

\bibitem{peng2018improving}
Xuefeng Peng, Yi~Ding, David Wihl, Omer Gottesman, Matthieu Komorowski,
  Li-wei~H Lehman, Andrew Ross, Aldo Faisal, and Finale Doshi-Velez.
\newblock Improving sepsis treatment strategies by combining deep and
  kernel-based reinforcement learning.
\newblock In {\em AMIA Annual Symposium Proceedings}, volume 2018, page 887.
  American Medical Informatics Association, 2018.

\bibitem{li2019optimizing}
Luchen Li, Matthieu Komorowski, and Aldo~A Faisal.
\newblock Optimizing sequential medical treatments with auto-encoding heuristic
  search in pomdps.
\newblock {\em arXiv preprint arXiv:1905.07465}, 2019.

\bibitem{killian2020empirical}
Taylor~W Killian, Haoran Zhang, Jayakumar Subramanian, Mehdi Fatemi, and
  Marzyeh Ghassemi.
\newblock An empirical study of representation learning for reinforcement
  learning in healthcare.
\newblock In {\em Machine Learning for Health}, pages 139--160. PMLR, 2020.

\bibitem{mnih2015human}
Volodymyr Mnih, Koray Kavukcuoglu, David Silver, Andrei~A Rusu, Joel Veness,
  Marc~G Bellemare, Alex Graves, Martin Riedmiller, Andreas~K Fidjeland, Georg
  Ostrovski, et~al.
\newblock Human-level control through deep reinforcement learning.
\newblock {\em nature}, 518(7540):529--533, 2015.

\bibitem{silver2016mastering}
David Silver, Aja Huang, Chris~J Maddison, Arthur Guez, Laurent Sifre, George
  Van Den~Driessche, Julian Schrittwieser, Ioannis Antonoglou, Veda
  Panneershelvam, Marc Lanctot, et~al.
\newblock Mastering the game of go with deep neural networks and tree search.
\newblock {\em nature}, 529(7587):484--489, 2016.

\bibitem{fuchs2020super}
Florian Fuchs, Yunlong Song, Elia Kaufmann, Davide Scaramuzza, and Peter Duerr.
\newblock Super-human performance in gran turismo sport using deep
  reinforcement learning.
\newblock {\em arXiv preprint arXiv:2008.07971}, 2020.

\bibitem{liu2020reinforcement}
Siqi Liu, Kay~Choong See, Kee~Yuan Ngiam, Leo~Anthony Celi, Xingzhi Sun, and
  Mengling Feng.
\newblock Reinforcement learning for clinical decision support in critical
  care: comprehensive review.
\newblock {\em Journal of medical Internet research}, 22(7):e18477, 2020.

\bibitem{yu2019reinforcement}
Chao Yu, Jiming Liu, and Shamim Nemati.
\newblock Reinforcement learning in healthcare: A survey.
\newblock {\em arXiv preprint arXiv:1908.08796}, 2019.

\bibitem{gottesman2019guidelines}
Omer Gottesman, Fredrik Johansson, Matthieu Komorowski, Aldo Faisal, David
  Sontag, Finale Doshi-Velez, and Leo~Anthony Celi.
\newblock Guidelines for reinforcement learning in healthcare.
\newblock {\em Nature medicine}, 25(1):16--18, 2019.

\bibitem{lange2012batch}
Sascha Lange, Thomas Gabel, and Martin Riedmiller.
\newblock Batch reinforcement learning.
\newblock In {\em Reinforcement learning}, pages 45--73. Springer, 2012.

\bibitem{fujimoto2019off}
Scott Fujimoto, David Meger, and Doina Precup.
\newblock Off-policy deep reinforcement learning without exploration.
\newblock In {\em International Conference on Machine Learning}, pages
  2052--2062, 2019.

\bibitem{osti_1561669}
Edmon Begoli, Tanmoy Bhattacharya, and Dimitri~F. Kusnezov.
\newblock The need for uncertainty quantification in machine-assisted medical
  decision making.
\newblock {\em Nature Machine Intelligence (Online)}, 1(1), 1 2019.

\bibitem{bellman1965dynamic}
Richard Bellman and Robert~E Kalaba.
\newblock {\em Dynamic programming and modern control theory}, volume~81.
\newblock Citeseer, 1965.

\bibitem{pmlr-v70-bellemare17a}
Marc~G. Bellemare, Will Dabney, and R{\'e}mi Munos.
\newblock A distributional perspective on reinforcement learning.
\newblock In Doina Precup and Yee~Whye Teh, editors, {\em Proceedings of the
  34th International Conference on Machine Learning}, volume~70 of {\em
  Proceedings of Machine Learning Research}, pages 449--458, International
  Convention Centre, Sydney, Australia, 06--11 Aug 2017. PMLR.

\bibitem{pmlr-v84-rowland18a}
Mark Rowland, Marc Bellemare, Will Dabney, Remi Munos, and Yee~Whye Teh.
\newblock An analysis of categorical distributional reinforcement learning.
\newblock In Amos Storkey and Fernando Perez-Cruz, editors, {\em Proceedings of
  the Twenty-First International Conference on Artificial Intelligence and
  Statistics}, volume~84 of {\em Proceedings of Machine Learning Research},
  pages 29--37, Playa Blanca, Lanzarote, Canary Islands, 09--11 Apr 2018. PMLR.

\bibitem{barth2018distributed}
Gabriel Barth-Maron, Matthew~W Hoffman, David Budden, Will Dabney, Dan Horgan,
  Dhruva Tb, Alistair Muldal, Nicolas Heess, and Timothy Lillicrap.
\newblock Distributed distributional deterministic policy gradients.
\newblock {\em arXiv preprint arXiv:1804.08617}, 2018.

\bibitem{agarwal2020optimistic}
Rishabh Agarwal, Dale Schuurmans, and Mohammad Norouzi.
\newblock An optimistic perspective on offline reinforcement learning.
\newblock In {\em International Conference on Machine Learning}, 2020.

\bibitem{caldeira2020deeply}
Jo{\~a}o Caldeira and Brian Nord.
\newblock Deeply uncertain: comparing methods of uncertainty quantification in
  deep learning algorithms.
\newblock {\em Machine Learning: Science and Technology}, 2(1):015002, 2020.

\bibitem{molnar2020interpretable}
Christoph Molnar.
\newblock {\em Interpretable machine learning}.
\newblock Lulu. com, 2020.

\bibitem{dabney2018distributional}
Will Dabney, Mark Rowland, Marc Bellemare, and R{\'e}mi Munos.
\newblock Distributional reinforcement learning with quantile regression.
\newblock In {\em Proceedings of the AAAI Conference on Artificial
  Intelligence}, volume~32, 2018.

\bibitem{foulon2018hemodynamic}
Pierre Foulon and Daniel De~Backer.
\newblock The hemodynamic effects of norepinephrine: far more than an increase
  in blood pressure!
\newblock {\em Annals of translational medicine}, 6(Suppl 1), 2018.

\bibitem{gottesman2018evaluating}
Omer Gottesman, Fredrik Johansson, Joshua Meier, Jack Dent, Donghun Lee,
  Srivatsan Srinivasan, Linying Zhang, Yi~Ding, David Wihl, Xuefeng Peng,
  et~al.
\newblock Evaluating reinforcement learning algorithms in observational health
  settings.
\newblock {\em arXiv preprint arXiv:1805.12298}, 2018.

\bibitem{rivera2020guidelines}
Samantha~Cruz Rivera, Xiaoxuan Liu, An-Wen Chan, Alastair~K Denniston, and
  Melanie~J Calvert.
\newblock Guidelines for clinical trial protocols for interventions involving
  artificial intelligence: the spirit-ai extension.
\newblock {\em bmj}, 370, 2020.

\bibitem{Clermont2004}
Gilles Clermont, John Bartels, Rukmini Kumar, Greg Constantine, Yoram Vodovotz,
  and Carson Chow.
\newblock In silico design of clinical trials: A method coming of age.
\newblock {\em Critical Care Medicine}, 32, 2004.

\bibitem{Urpi21}
Núria~Armengol Urpí, Sebastian Curi, and Andreas Krause.
\newblock Risk-averse offline reinforcement learning, 2021.

\bibitem{johnson2018comparative}
Alistair~EW Johnson, Jerome Aboab, Jesse~D Raffa, Tom~J Pollard, Rodrigo~O
  Deliberato, Leo~A Celi, and David~J Stone.
\newblock A comparative analysis of sepsis identification methods in an
  electronic database.
\newblock {\em Critical care medicine}, 46(4):494--499, 2018.

\bibitem{mimiciii}
Alistair~EW Johnson, Tom~J Pollard, Lu~Shen, Li{-}wei~H Lehman, Mengling Feng,
  Mohammad Ghassemi, Benjamin Moody, Peter Szolovits, Leo~Anthony Celi, and
  Roger~G Mark.
\newblock Mimic-iii, a freely accessible critical care database.
\newblock {\em Scientific data}, 3:160035, 2016.

\bibitem{mimiciiidata}
Alistair~EW Pollard, Tom J abd~Johnson.
\newblock The mimic-iii clinical database.
\newblock \url{http://dx.doi.org/10.13026/C2XW26}, 2016.

\bibitem{cho2014learning}
Kyunghyun Cho, Bart Van~Merri{\"e}nboer, Caglar Gulcehre, Dzmitry Bahdanau,
  Fethi Bougares, Holger Schwenk, and Yoshua Bengio.
\newblock Learning phrase representations using rnn encoder-decoder for
  statistical machine translation.
\newblock {\em arXiv preprint arXiv:1406.1078}, 2014.

\bibitem{bighamian2013analytic}
Ramin Bighamian, Andrew~T Reisner, and Jin-Oh Hahn.
\newblock An analytic tool for prediction of hemodynamic responses to
  vasopressors.
\newblock {\em IEEE Transactions on Biomedical Engineering}, 61(1):109--118,
  2013.

\bibitem{bighamian2014prediction}
Ramin Bighamian, Sadaf Soleymani, Andrew~T Reisner, Istvan Seri, and Jin-Oh
  Hahn.
\newblock Prediction of hemodynamic response to epinephrine via model-based
  system identification.
\newblock {\em IEEE journal of biomedical and health informatics},
  20(1):416--423, 2014.

\bibitem{chan2016listen}
William Chan, Navdeep Jaitly, Quoc Le, and Oriol Vinyals.
\newblock Listen, attend and spell: A neural network for large vocabulary
  conversational speech recognition.
\newblock In {\em 2016 IEEE International Conference on Acoustics, Speech and
  Signal Processing (ICASSP)}, pages 4960--4964. IEEE, 2016.

\bibitem{schaul2015prioritized}
Tom Schaul, John Quan, Ioannis Antonoglou, and David Silver.
\newblock Prioritized experience replay.
\newblock {\em arXiv preprint arXiv:1511.05952}, 2015.

\bibitem{vapnik1992principles}
Vladimir Vapnik.
\newblock Principles of risk minimization for learning theory.
\newblock In {\em Advances in neural information processing systems}, pages
  831--838, 1992.

\bibitem{Zhang2010StructuralRM}
X.~Zhang.
\newblock Structural risk minimization.
\newblock In {\em Encyclopedia of Machine Learning}, 2010.

\bibitem{kullback1951information}
Solomon Kullback and Richard~A Leibler.
\newblock On information and sufficiency.
\newblock {\em The annals of mathematical statistics}, 22(1):79--86, 1951.

\bibitem{kingma2014adam}
Diederik~P Kingma and Jimmy Ba.
\newblock Adam: A method for stochastic optimization.
\newblock {\em arXiv preprint arXiv:1412.6980}, 2014.

\end{thebibliography}

\newpage
\section*{Supplementary Information}
\subsection*{Appendix A: Cohort Details}
Our total patient cohort consists of 18,472 patients, out of which 1,828 were non-survivors.

\begin{table}[h]
    \centering
    \caption{Cohort Details}
    \begin{tabular}{lllll}
        \toprule
         
         \multirow{2}{*}[-1em]{Cohort} \\ 
        \addlinespace[3pt]
        {} &  \% Female &Mean Age  & Mean ICU Stay &Total Population \\
        \midrule
        Overall & 42.33 \% & 66.05 & 7 days 15 hours & 18472 \\
        Non-Survivors & 42.67 \% & 68.8 & 9 days 13 hours & 1828 \\
        Survivors & 42.14 \% & 65.91 & 5 days 13 hours & 16644 \\
        \bottomrule
    \end{tabular}
    
\end{table}

This resulted in an experience replay consisting a total of 2596604 transitions.

\subsection*{Appendix B: Neural Network Architectures and Implementation Details}
\subsubsection*{Physiology-driven Autoendcoder}
\textbf{Encoder :}
\begin{itemize}
\item Patient Encoder: Multi-layer feed-forward neural network, with 3 hidden layers with 64 nodes each, followed by exponential linear unit, (eLU) non-linearity applied element wise.

\item Transition: Multi-layer feed-forward neural network, with 8 hidden layers with 128 nodes each, followed by exponential linear unit, (eLU) non-linearity applied element wise.

\item RNN: Gated recurrent unit, based RNN, with 1 hidden layer, with 64 nodes.
\end{itemize}
We note that, as inputs for the network specifically the RNN, we included all vitals, and SOFA-related scores,  including the four dimensional observations, systolic blood pressure, diastolic blood pressure, mean blood pressure and heart rate.

For training, we used Adam \cite{kingma2014adam}, with a low learning rate (1e-5), the corruption was only introduced after the model has been trained for several epochs. For RL representation we used the model trained with 10\% corruption.



\subsubsection*{Denoising Lab Autoencoder}
This is comprised of three GRU networks stacked on top of each other.

\begin{itemize}
\item Network 1 : 12 hidden units, with 512 nodes, outputs a 128 node vector.

\item Network 2 : 5 hidden units 128 nodes each, outputs a 10 dimensional vector.

\item Network 3: 3 hidden units, with 10 nodes each, the last of which is taken as our hidden lab representation.
\end{itemize}

We train this again using Adam, and corruption is gradually introduced starting from 0\% to 50\%. We use the network trained under 50\% corrupted inputs, when inferring the hidden lab representation for RL.

We standardized all the labs before feeding into the network.

\subsubsection*{Imitation Learning}
Muli-layer neural network with 4 hidden layers: 3 with node size 512, and the last 256. All hidden (and input) layers are followed by a rectified linear unit (reLU) non-linearity.

Training was again using Adam with a standard learning rate, and we minimized a negative log-likelihood loss, which is standard in classification problems.

\subsubsection*{Bootstrapping and Deep ensembles}
To learn each bootstrapped network, we first sampled from the all patients to arrive at a bootstrapped patient list. Then we train the networks, for 2 or 3 epochs each (to have further randomness), using a process identical to training the main RL algorithm.

For the uncertainty quantification step, we trained the majority of bootstrapped ensembles on as little as 40\% of total patients, however for the results presented under vasopressor administration, we only considered ensembles which were trained on a cohort of 65-80\% of patients. The number of patients was also picked at random. There were 20 such bootstrapped ensembles.

\subsubsection*{Distributional Q learning}

We use the standard C51 training algorithm as in \cite{pmlr-v70-bellemare17a}. Q network was a multi-layer neural network. Apart from the weighted sampling described in the main body of the text, training steps were all standard. 

We  use a target network, and update the target networks using polyak target updating with $\tau=0.005$. (i.e. after every iteration/training step we set the target network weights to a linear combination of it's own weights, weighted by (1-$\tau$) an the Q network weights, weighted by $\tau$). This kind of target network is common amongst all deep Q learning, algorithms.

We summarize the hyper-parameters involved in table below. 

\begin{table}[h]
    \centering
    \caption{RL algorithm hyper-parameters}
    \begin{tabular}{ll}
        \toprule
         
         \multirow{2}{*}[-1em]{Hyper-parameter} \\ 
        \addlinespace[3pt]
        {} &  Value \  \\
        \midrule
        Support size & 51 \\
        Maximum value &  18 \\
        Minimum value &  -18 \\
        $\gamma$ & 0.999\\
        Batch size &  100 \\
        Number of iterations &  51932 \\
        Optimizer & Adam\\
        Learning rate & $3*10^{-4}$\\
        $\tau$ & 0.005\\
        \bottomrule
    
    \end{tabular}
    
\end{table}

\subsection*{Appendix C: Additional Results}
\subsubsection*{Representation Learning}
The physiology-driven autoencoder was capable of almost perfect reconstruction of the input when it was trained, and evaluated without corruption. The unnormalized mean square loss per time-step, on the 4 dimensional output was around 6.

With 10\% corruption probability, the loss --- evaluated also with corruption, was around 45 and, with 25\% corruption this went up to close to 60. We experimented with 50\% corruption rates, but however results were far less impressive as indicated in the main text.

\subsubsection*{RL Results}
\begin{figure}
    \centering
    \includegraphics[scale=0.35]{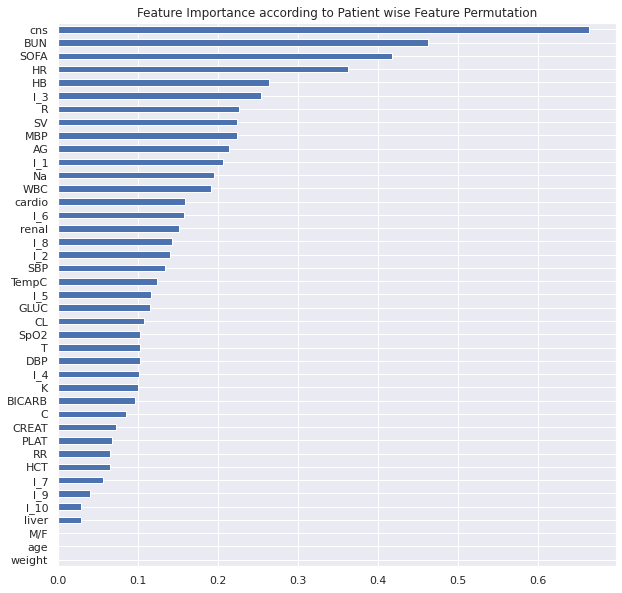}
    \caption{Feature Importance measured by feature permutation}
    \label{fig:imps}
\end{figure}

In this section, we present further results of the distributional RL algorithm, and uncertainty quantification. First, in Figure \ref{fig:imps} we present the feature importance of all features.

\begin{figure}[h!]

\includegraphics[width=150 pt]{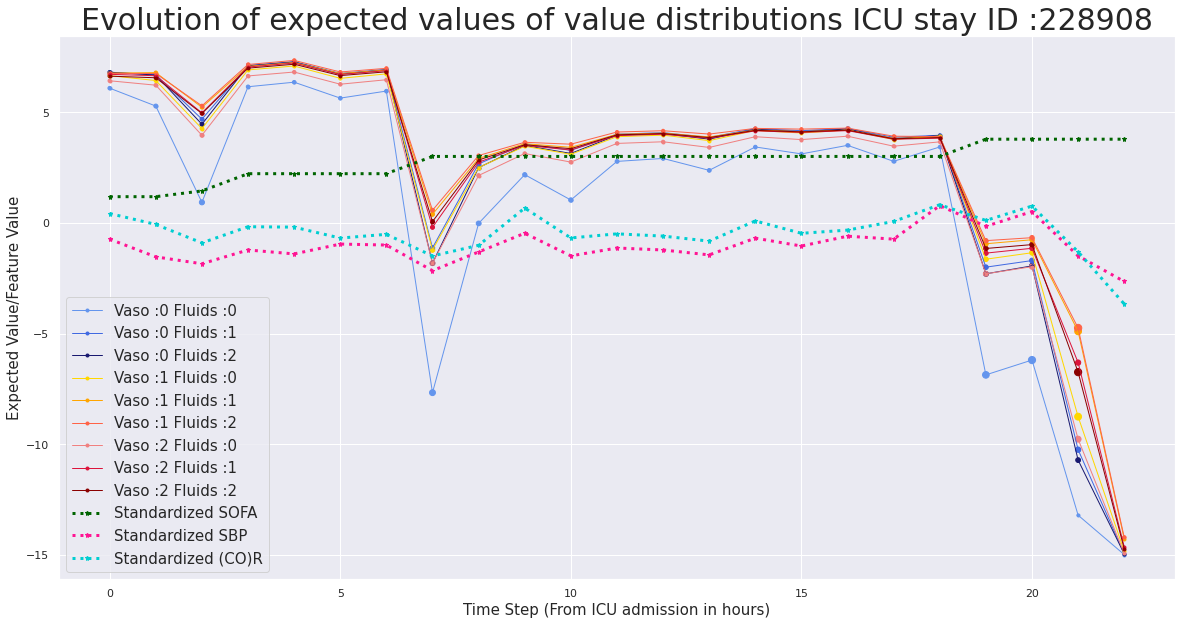}
\includegraphics[width=150 pt]{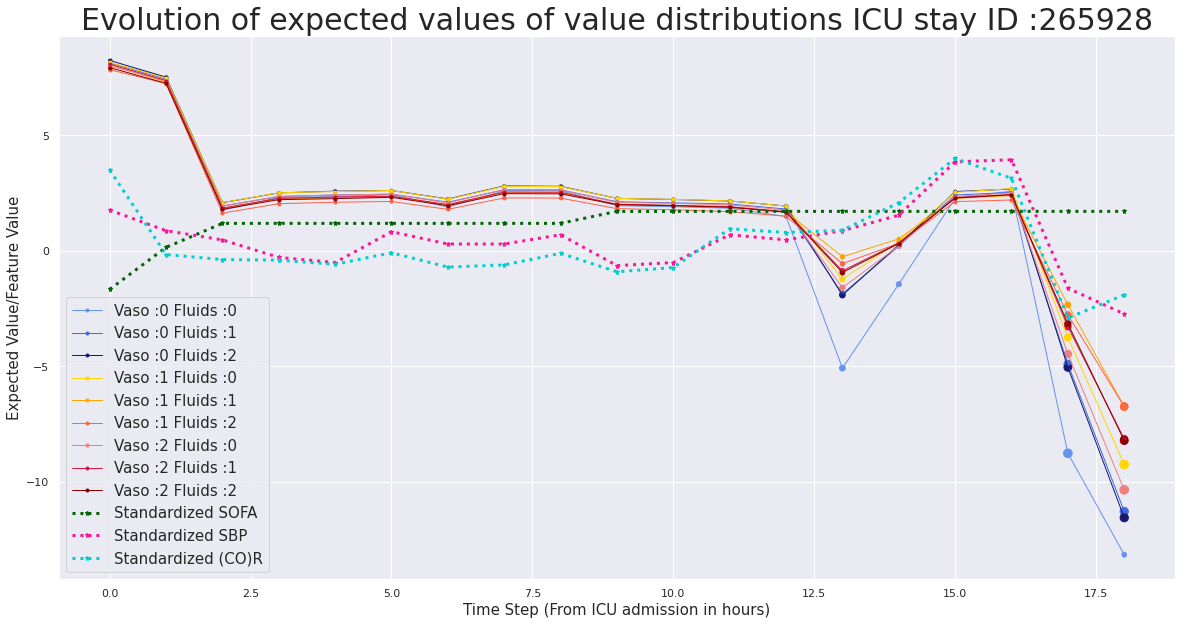}
\includegraphics[width=150 pt]{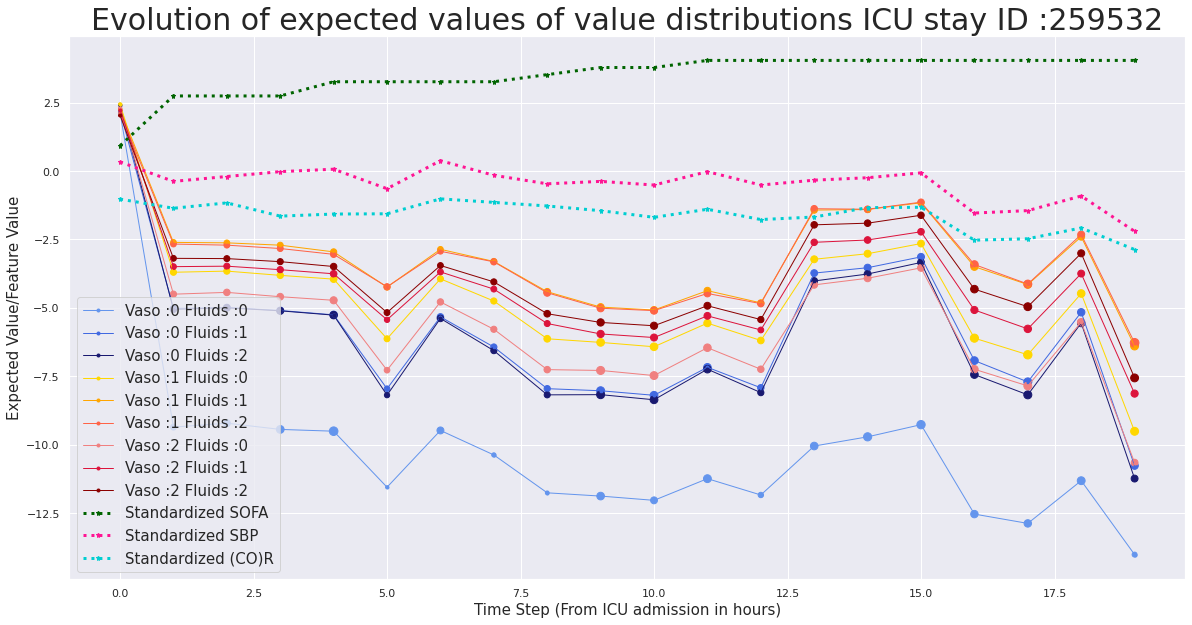}

\includegraphics[width=150 pt]{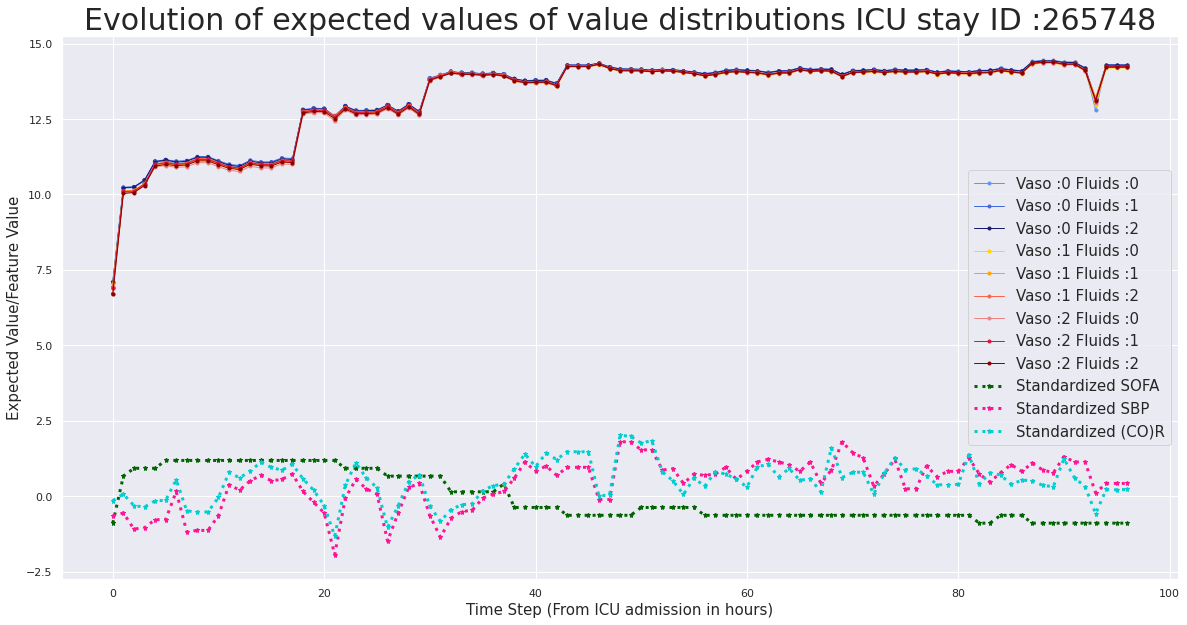}
\includegraphics[width=150 pt]{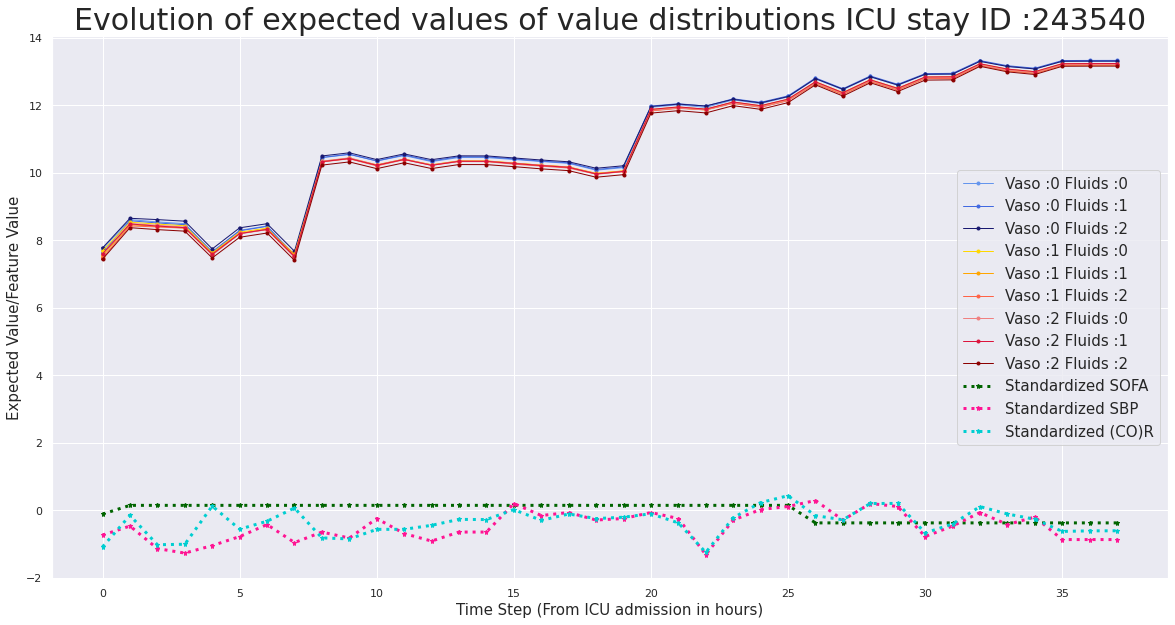}
\includegraphics[width=150 pt]{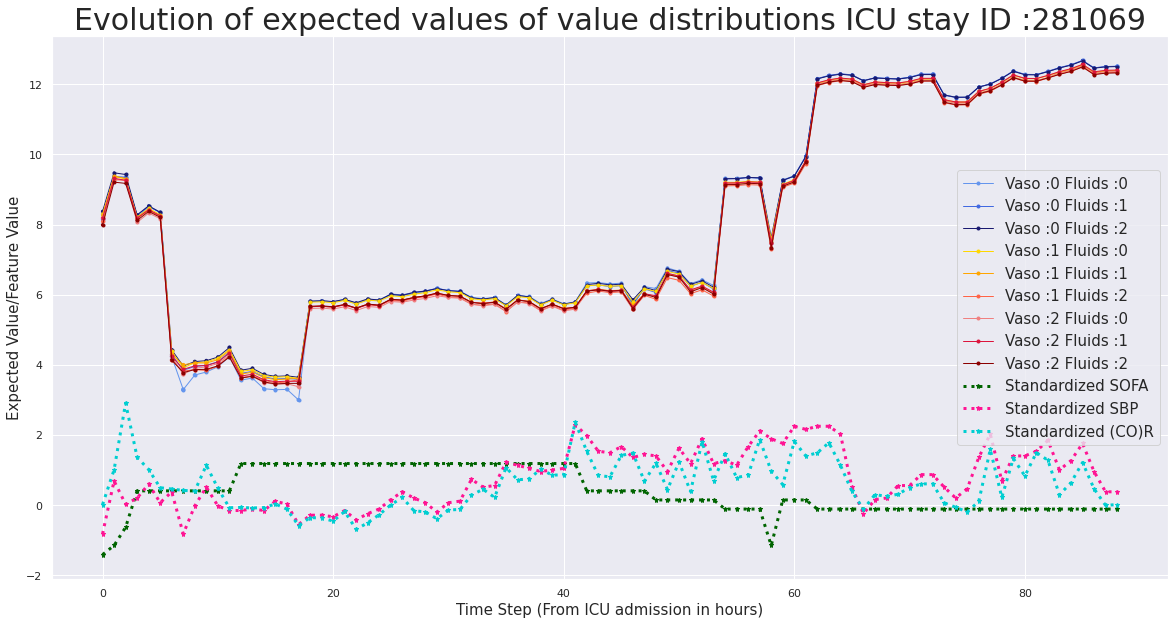}

\caption{Expected Values of random validation patients, \textbf{Top}: Non-survivors, \textbf{Bottom}: Survivors. As with Figure 4, the blob size indicate the uncertainty}
\label{fig:val_exps}
\end{figure}

Figure \ref{fig:val_exps} shows, expected value trajectories, of randomly selected validation patients. As we have mentioned previously these results indicate the generalizability of our value networks.
 
To be consistent with previous work (e.x. \cite{raghu2017deep}), we present heat plots of global actions, overall, low SOFA ($<5$), medium SOFA ($\ge 5 <15$) and high SOFA ($>15)$ separately. We further provide last 24 hours on non-survivors and results from decisions taken with respect to the expected value of the an ensembled weighted distribution, (corresponding to $\beta=1,\lambda=0$) and  $\beta=0.8,\lambda=0.25$.

However, at its core, our approach strives to extract and use patient specific recurrent representations to learn personalized treatments. Therefore global analysis is unlikely to provide much insight into the intricacies that underline the decision making process. Further when analyzing the proposed treatment, it should be noted that each action is proposed considering only the current, actual state. Therefore for a fixed patient trajectory at a fixed time, the agent does not know what it has proposed previously, nor how its action would have impacted the state. 

\begin{figure}[h!]
    \centering
    \includegraphics[scale=0.15]{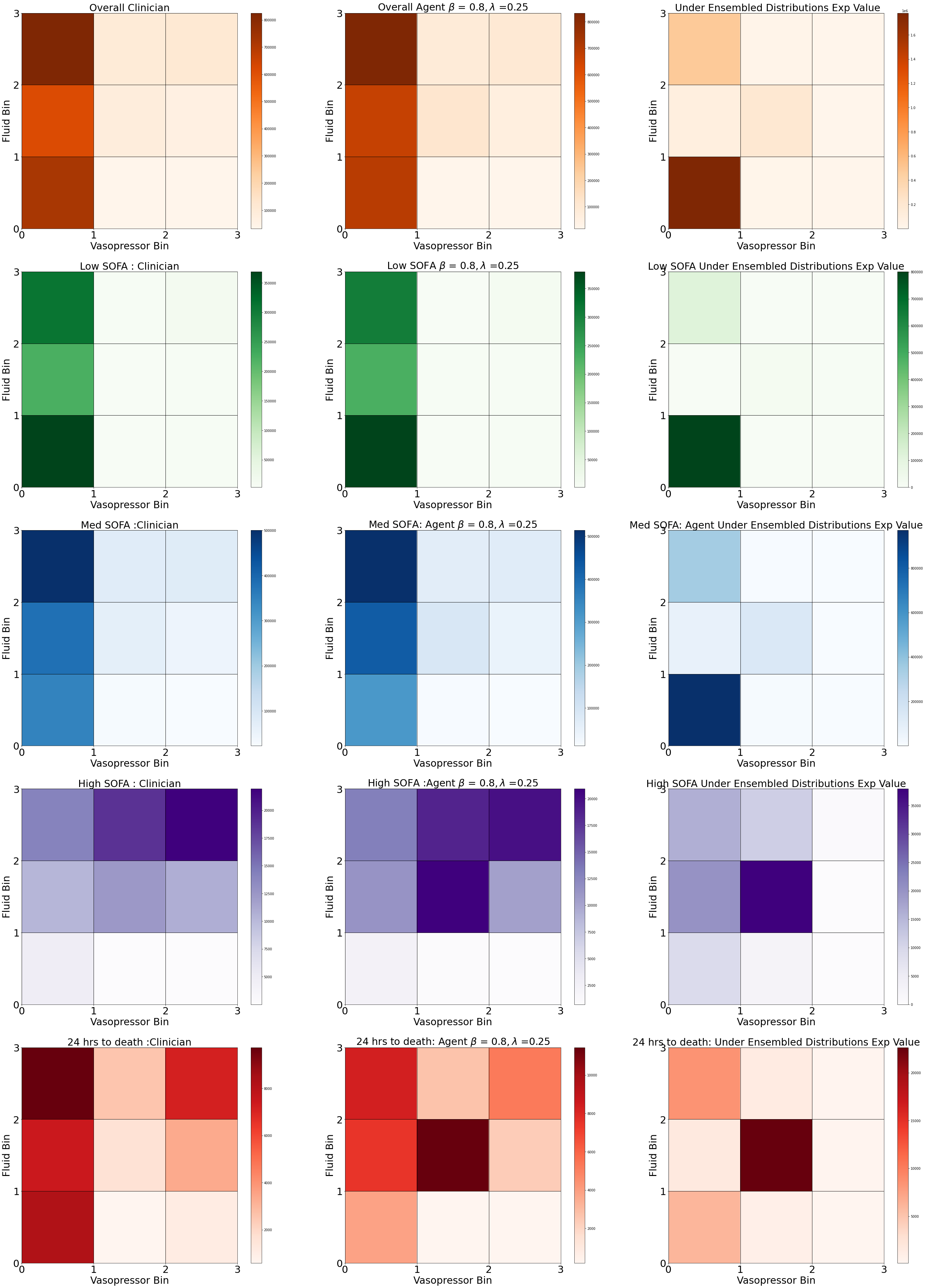}

    \label{fig:uncertain_acts}
    
      \caption{\textbf{L} : Heat-plots for recommended actions, under $\beta=0.8$, $l\lambda=0.25$ and Ensembled Distribution Expected Values. Shown are clinician's vs Agent for overall (orange), low sofa (green), medium sofa (blue), high score (purple) and non survivors last 24 hrs (red).}

\end{figure}

The most striking difference is for non survivors near death states. Our methods consistently recommend vasopressors. 
It is also interesting that RL methods have in general preferred low/medium (corresponding to 1) vasopressors and fluids as opposed to high doses (2). Just as we mentioned in the main text, when ensembled, agents do not recommend fluids for survivors' less critical states. 
It must be noted however that the agent trained on the whole cohort did have fluids recommended regularly, but there is disagreement amongst the ensembles.

\subsubsection*{Uncertainty Quantification Results}

In this section, we briefly mention results of uncertainty quantification.

\begin{figure}[t]
\includegraphics[width=200 pt]{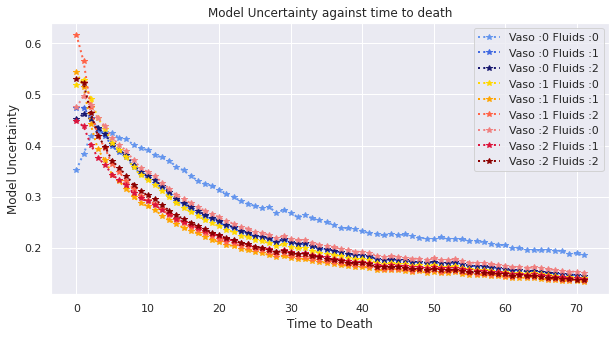}
\hspace{10pt}
\includegraphics[width=200 pt]{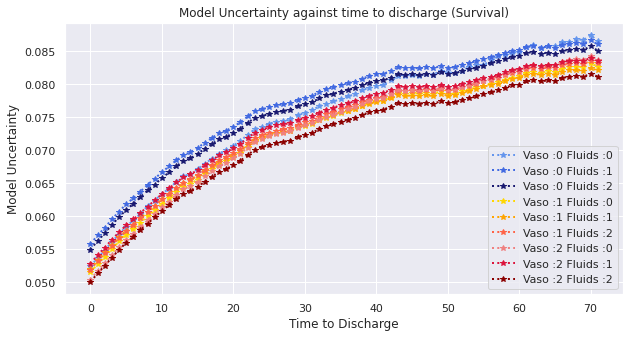}
\caption{\textbf{Left :} Model Uncertainty with time to death for non-survivors, \textbf{Right :} Model Uncertainty with time to discharge for survivors}
\label{fig:uncertainty}
\end{figure}

The common pattern is that for most patients who have died, the model is less confident about its value distributions as they become closer to death. Uncertainty among each action varies from patient to patient. However for survivors this behavior is the exact opposite, as the agent is more confident of its results and becomes even more  confident as the patient gets closer to discharge from the ICU. We illustrate this in Figure \ref{fig:uncertainty}, which presents averaged model uncertainty with time to death and discharge for non-survivors and survivors respectively.

This observation is not surprising since death states are relatively uncommon, and also there are a wide variety of ways a septic patient may face increased mortality risk. However for survivors, we do expect all of them to approach a \textit{healthy} state as they approach eventual discharge.

Table 3 presents average uncertainty, among all patient states, grouped by the training and validation datasets and whether the patient was a survivor or a non-survivor. As we mentioned in the main text, the uncertainty is much higher for non survivors than survivors. Further uncertainties for validation non-survivors are higher than training non-survivors. However for survivors the training and validation uncertainty are very similar on average.

Both Table 3 and Figure \ref{fig:uncertainty} agree with our expectations, because near-death states, are relatively uncommon, and also there could be a lot of different ways a septic patient may have increased mortality risk. However, for survivors, we do expect our agent to be confident of their survival, as their states should approach a \textit{healthy} state.

\begin{table}[t]
    \centering
    \caption{Mean Model Uncertainty for Survivors and Non-Survivors in training and validation datasets} 
    
    \begin{tabular}{lllll}
        \toprule
         
         \multirow{2}{*}[-1em]{Action/Cohort} \\ 
        \addlinespace[3pt]
        {} &  Non-Survivors Train.  &  Survivors Train. & 
        Non Survivors Val.  &   Survivors Val. 
        \\
        \midrule
        Vaso 0 Fluids 0 & 0.1916 & 0.0887 & 0.3617 & 0.0861 \\
        Vaso 0 Fluids 1 & 0.1591 & 0.0855 & 0.3085 & 0.0894 \\
        Vaso 0 Fluids 2 & 0.1587 & 0.0846 & 0.3104 & 0.0879 \\
        Vaso 1 Fluids 0 & 0.1547 & 0.0820 & 0.3066 & 0.0850 \\
        Vaso 1 Fluids 1 & 0.1451 & 0.0815 & 0.2676 & 0.0847 \\
        Vaso 1 Fluids 2 & 0.1482 & 0.0827 & 0.2776 & 0.0850 \\
        Vaso 1 Fluids 0 & 0.1634 & 0.0839 & 0.3135 & 0.0853 \\
        Vaso 2 Fluids 1 &  0.1498 & 0.0831 & 0.2710 & 0.0860\\
        Vaso 2 Fluids 2 &  0.1488 & 0.0808 & 0.2850 & 0.0832 \\
      
        \bottomrule
        
    \end{tabular}

\end{table}

\subsection*{Appendix D: Limitations and Open Problems}   

As stated in the main text, and discussed in previous work, the main limitation of \emph{any} data driven or computational approach to finding optimal treatment is proper evaluation of the learned policy. In this work we relied on medical expertise and physiologic knowledge in interpreting the results, but evaluating learned policies is an active research area in offline RL, and future research could find better methods which are more suited to critical care applications. 

\begin{figure}[h!]

\includegraphics[width=200 pt]{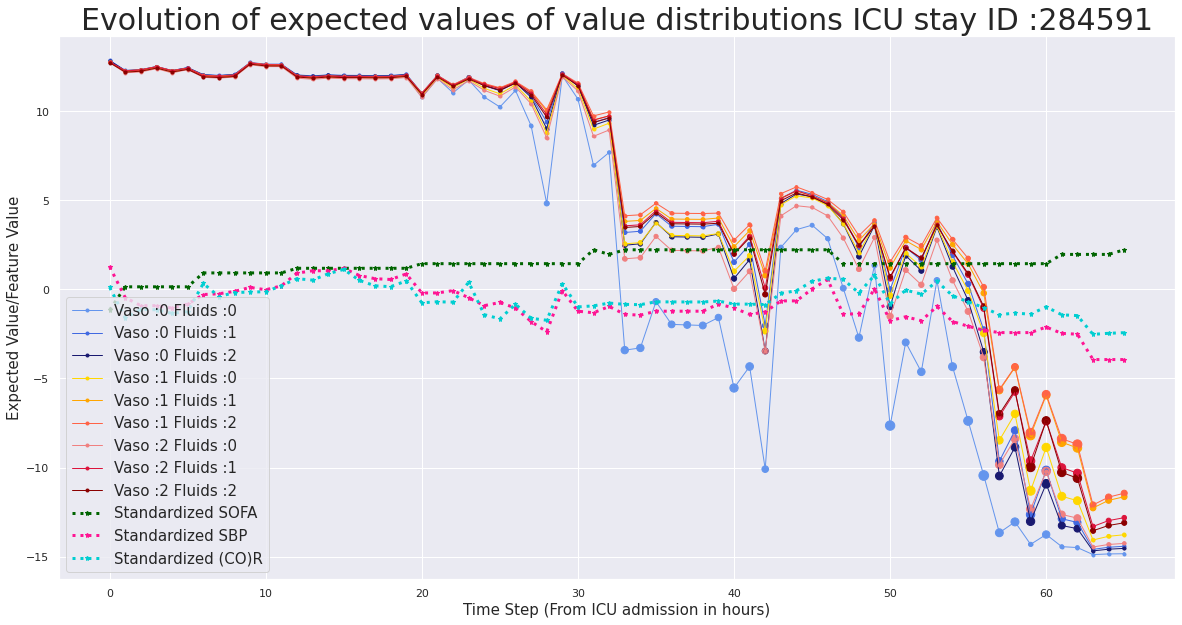}
\hspace{10pt}
\includegraphics[width=200 pt]{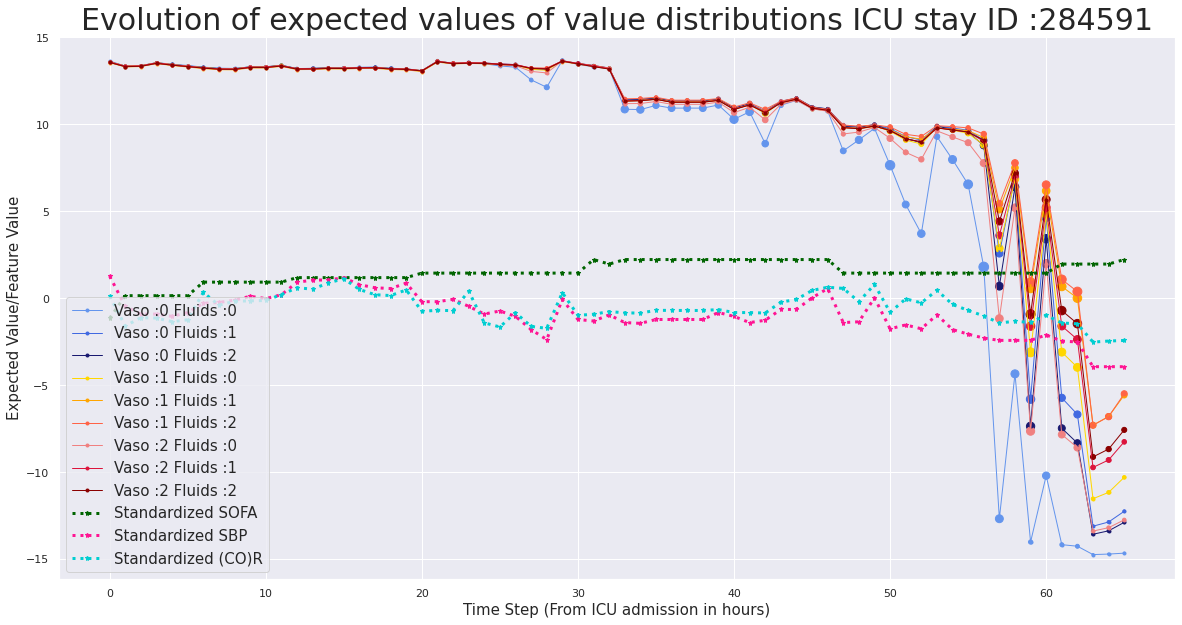}

\caption{Expected Values of non-survivor, \textbf{Left}: Trained for 2 epochs, \textbf{Right} : Trained for 7 epochs}
\label{fig:overfit}
\end{figure}

A related issue is model selection. Like supervised learning, it has been shown previously that training deep RL algorithms longer on the same dataset can result in poor performance and overfitting. A lack of an obvious evaluation metric (such as test accuracy for a classification problem) makes model selection complicated. We used results after only two full passes of the dataset (51932 iterations), observing that the results don't make the same sense, clinically, when it is trained for too long. Indeed figure \ref{fig:overfit} shows the expected value evolution for a validation cohort non-survivor for different weights. As we can see, its results are far too optimistic when the patient is a few hours away from death, if trained longer. However, the vasopressor recommendation results for non-survivors, which was presented earlier, do hold for all the different training weights.

In the context of sepsis treatment, as we mentioned under Discussion, a further challenge is designing rewards. For example even \emph{survivors} have a high risk of relapse, and their physiologic age is significantly higher than their actual age. Therefore it can be argued that survival at the ICU should not be \emph{rewarded} as much as (in absolute value) death. Further organ damage and mortality could be competing objectives for some patients. Whilst it is possible to have a weighted combination, of both as we did, a multi-objective RL framework could also be looked into. As we mentioned before we hope to explore these questions in future work.

\subsubsection{Future Work}
There are other avenues we would like to explore. 

\textbf{Model-based RL with  physiological models}: Model-based RL aims to explicitly model the underlying environment and then use this information in various ways for control. \footnote{It could in theory be argued that our work itself is a model based and model free hybrid method.} This paradigm provides a natural place to incorporate mechanistic models, which could potentially help both control and interpretability. Clearly, the availability of more granular data, or of additional domains of data, could allow better estimation of the underlying physiological model and thus reduce uncertainty. 

\textbf{Reward Structure}: Our reward structure was based on previous work and has clinical appeal. However, rewards are an essential component of any RL algorithm and is the only place where the agent can judge the merit of its proposed actions. This is potentially another place to include physiological knowledge. Ideally, we would want our reward structure to capture an accurate mortality risk, and an organ damage score, with each state. Risk-based rewards, rooted in anticipated evolution over a meaningful clinical horizon, should be considered in future schemes. 

\textbf{Risk Averse RL}: It could be argued that, maximizing the sum of expected future rewards may not best reflect the end goals of safety critical domains. Whist the rewards can be engineered to promote risk aversion, risk averse RL is a fast growing research area, which we are keen to explore, if the RL objective itself can be tweaked to be more suitable for critical care research.

\end{document}